\documentclass[manuscript]{acmart}
\AtBeginDocument{%
  }

\setcopyright{acmlicensed}
\copyrightyear{2026}
\acmYear{2026}
\acmDOI{XXXXXXX.XXXXXXX}
\acmConference[XXX'26]{XXX}{XXXX XX--XX,
  2026}{XXXX, XX}
\acmISBN{978-1-4503-XXXX-X/2026/06}

\usepackage{graphicx}
\usepackage{booktabs}
\usepackage{multirow,array}
\usepackage[table]{xcolor}
\usepackage{enumitem}
\usepackage{float}
\usepackage{amsmath}
\usepackage{amsfonts}
\usepackage{algorithm}
\usepackage{algorithmic}
\usepackage{tabularx}
\usepackage{color, xcolor}
\usepackage{caption}
\usepackage{threeparttable}
\usepackage{makecell}
\begin{document}

\title{Fine-grained Verification via Diagnostic Reasoning Supervision for Aspect Sentiment Triplet Extraction}

\author{Wenna Lai}
\affiliation{%
  \institution{\centering{The Hong Kong Polytechnic University}}
  \city{Hong Kong SAR}
  \country{China}
}
\email{winnelai05@gmail.com}

\author{Haoran Xie}
\authornote{Corresponding author}
\affiliation{%
  \institution{Lingnan University}
  \city{Hong Kong SAR}
  \country{China}
  }
\email{hrxie@ln.edu.hk}

\author{Guandong Xu}
\affiliation{%
  \institution{\centering{Education University of Hong Kong}}
  \city{Hong Kong SAR}
  \country{China}
}
\email{gdxu@eduhk.hk}

\author{Qing Li}
\affiliation{%
   \institution{\centering{The Hong Kong Polytechnic University}}
  \city{Hong Kong SAR}
  \country{China}
  }
\email{qing-prof.li@polyu.edu.hk}

\author{S. Joe Qin}
\affiliation{%
   \institution{Lingnan University}
  \city{Hong Kong SAR}
  \country{China}
  }
\email{joeqin@ln.edu.hk}

\renewcommand{\shortauthors}{Lai et al.}


\begin{abstract}
Aspect Sentiment Triplet Extraction (ASTE) aims to identify aspect terms, opinion terms, and sentiment polarities as structured triplets, providing essential inputs for downstream information system applications such as opinion mining, explainable recommendations, and review summarization. Prior work mainly focuses on end-to-end extraction, while post hoc verification of extracted triplets remains comparatively underexplored. This gap limits the reliability of ASTE systems, since predicted triplets may be locally plausible while being globally invalid. Moreover, candidate invalidity is multi-faceted and candidate usability is inherently graded, motivating a fine-grained verification mechanism that can filter or re-rank outputs from diverse extractors.
In this paper, we propose FiVeD, a framework for \textbf{Fi}ne-grained \textbf{Ve}rification with \textbf{D}iagnostic reasoning supervision. Specifically, the verifier is trained with multiple complementary objectives, including validity classification and quality score estimation as primary tasks, with error type classification and rationale generation as auxiliary tasks. We define hierarchical error categories and construct plausible incorrect triplets under semantic and syntactic constraints, and leverage an off-the-shelf LLM with task-specific rubrics to produce quality scores and diagnostic rationales. During inference, the resulting quality scores are used to filter candidate outputs, supporting adjustable precision-recall tradeoffs. Experiments across multiple ASTE baselines demonstrate that FiVeD consistently improves extraction performance by up to $3.53$ F1 points as a plug-and-play verification module.

\end{abstract}



\maketitle

\section{Introduction}

Aspect-Based Sentiment Analysis (ABSA) is a fine-grained sentiment analysis task that aims to identify sentiment opinions toward specific aspects in a given text \cite{tkde/SchoutenF16, lai2025star}. As a representative compound task in ABSA, Aspect Sentiment Triplet Extraction (ASTE) further requires models to extract aspect terms, opinion terms, and the sentiment polarities expressed toward the aspects as structured triplets. Compared with sentence-level sentiment analysis or aspect-level sentiment classification, ASTE provides a more complete and interpretable representation of opinion information by jointly modeling multiple sentiment-related elements and their relations \cite{Peng_Xu_Bing_Huang_Lu_Si_2020, naglik-lango-2024-aste}. Such structured sentiment information is a critical input to downstream information system applications such as opinion mining, explainable recommendation, faceted product analysis, and review summarization, all of which require reliable information retrieval \cite{lai2025llmsteamup, Cui_Wang_Ho_Cambria_2023}.

Recent studies have made substantial progress in ASTE through end-to-end extraction models, which formulate the task from different perspectives, including sequence tagging \cite{yan-etal-2021-unified, xu-etal-2020-position}, machine reading comprehension (MRC)-based extraction \cite{Dual-MRC21, zhai-etal-2022-com, Zou2024AMS}, span-based extraction \cite{xu-etal-2021-learning, zhao-etal-2024-dual}, table filling \cite{SIMSTAR, sun-etal-2024-minicongts, IJCAI-TT-TABLE}, and generative modeling \cite{coling/GaoFLLLLBY22, acl/Zhang0DBL20}. 
Despite their methodological differences, these approaches are primarily optimized to extract accurate triplet predictions in an end-to-end manner. 
In practice, their inference often involves partially decoupled sub-decisions, such as element extraction, aspect-opinion pairing, and sentiment classification \cite{DGSEP, bodke-etal-2025-pastel, EPMEI}. 
Some methods further refine predictions within the extraction pipeline through simple post-processing steps, such as heuristic filtering or majority voting \cite{bodke-etal-2025-pastel, acl/MvP}. 
However, these steps typically rely on a constrained set of extraction signals, such as model probabilities, decoding scores, or prediction frequencies. 
Although such signals are useful for selecting final predictions, they provide only coarse evidence for candidate reliability and place limited emphasis on candidate validity, which concerns whether a predicted triplet is semantically supported as a complete structure. 
This limitation is further amplified when these signals are tied to specific model architectures or decoding procedures \cite{Li2024ASO}, which restricts their use as general indicators of triplet validity and leaves downstream applications with a limited basis for assessing prediction reliability.

\begin{figure}
\centering
\includegraphics[width=0.6\linewidth]{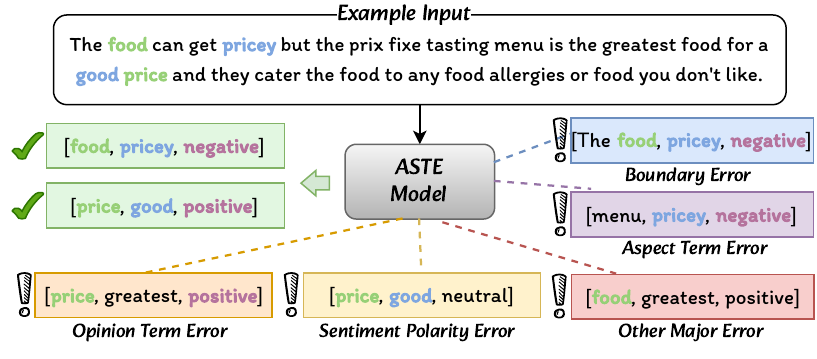}
\caption{Motivating example of fine-grained triplet verification. An ASTE model may produce both correct triplets and plausible invalid candidates. FiVeD aims to verify each candidate and diagnose its error type before final prediction.}
\label{fig:example}
\end{figure}

As illustrated in Figure~\ref{fig:example}, the validity of a triplet depends on the joint consistency of its constituent elements and their contextual relation, rather than on the plausibility of each component in isolation. 
Variants of a gold triplet may appear locally plausible while still exhibiting hierarchical errors, ranging from simple boundary errors to single element errors (e.g., opinion term error) and compound errors involving multiple triplet components. 
These cases are difficult to resolve solely through the extraction process and motivate a candidate verification mechanism that complements conventional ASTE extraction. 
Such a mechanism verifies each candidate triplet against the input sentence to assess whether it is semantically supported as a complete structure and to provide diagnostic signals about candidate usability. 
However, effective verification for ASTE remains challenging. 
First, candidate invalidity is fine-grained and structured, involving possible errors in span boundaries, triplet elements, element pairings, sentiment polarity, or multiple components simultaneously. 
Binary validity supervision is too coarse to characterize the nature and severity of candidate errors. 
Second, candidate usability is inherently graded, since candidates may be fully supported, partially supported, weakly supported, or contradicted by the input sentence. 
Without fine-grained verification, valid candidates with low extraction confidence may be suppressed, while candidates that appear locally plausible but are not semantically supported by the sentence may still be retained. 
Therefore, a verifier that produces graded and interpretable signals can better capture both triplet  validity and the quality of candidate support. 

To address these challenges, we propose \textbf{FiVeD}, a framework for \textbf{Fi}ne-grained \textbf{Ve}rification with \textbf{D}iagnostic Reasoning Supervision for reliable ASTE.
Instead of replacing existing extractors, FiVeD learns a plug-and-play verifier that assesses the reliability of candidate triplets produced by a base extractor, and operates in two stages. 
In the first stage, we construct verification supervision from gold ASTE annotations. 
Specifically, we perform counterfactual candidate mining to synthesize plausible invalid triplets under semantic and syntactic constraints, exposing the verifier to realistic extraction errors. 
We then assign hierarchical error types to each candidate based on common ASTE error patterns, providing fine-grained supervision beyond binary validity labels. 
In addition, we use an off-the-shelf LLM to derive quality scores and rationales, which together with hierarchical error types form diagnostic reasoning supervision for fine-grained verification.
In the second stage, we train the verifier with multi-task learning and automatically learned task weights. 
The verifier jointly learns validity classification and quality score estimation as main tasks, together with error type classification and rationale generation as auxiliary tasks. 
During inference, it estimates the validity and quality of candidate triplets and uses the predicted scores to rescore, filter, or rerank extractor outputs. 

Since FiVeD does not require modifying the base extractor, it is model-agnostic and can be flexibly integrated with diverse ASTE architectures while supporting adjustable trade-offs between precision and recall. 
In this work, we focus on improving generative extractors, which operate in a more open output space than sequence labeling, table filling, and span-based methods and can naturally produce diverse candidate triplets. 
These properties make them particularly suitable for post-hoc verification through filtering and reranking. 
We evaluate FiVeD across multiple generative ASTE extractors to assess its effectiveness and flexibility relative to state-of-the-art baselines.
Experimental results show that FiVeD consistently improves extraction performance across baselines. 
Further analysis demonstrates the contribution of its key components, highlighting the benefits of fine-grained verification with diagnostic reasoning supervision.

In summary, the main contributions of this paper are as follows:
\begin{itemize}
    \item We propose FiVeD, a fine-grained verification framework for reliable ASTE, which improves triplet validity assessment through quality estimation and diagnostic reasoning supervision, and can be flexibly integrated with diverse ASTE models as a plug-and-play verification module.

    \item We construct diagnostic reasoning supervision for ASTE verification by defining hierarchical error categories, generating plausible incorrect triplets under semantic and syntactic constraints, and leveraging an off-the-shelf LLM with task-specific rubrics to produce quality scores and diagnostic rationales.

    \item We conduct experiments on four benchmark datasets across across multiple ASTE baselines, showing that our framework consistently improves extraction performance by up to $3.53$ F1 points and enables adjustable precision-recall control through quality score thresholding.
\end{itemize}

The remainder of this paper is organized as follows. Section~\ref{sec:method} introduces the proposed FiVeD framework, which is evaluated in Section~\ref{experiment}. Section~\ref{sec:discussion} provides further analyses of task weighting and quality score supervision across different error types. Section~\ref{sec:relatedwork} reviews recent progress in ASTE and verification techniques for structured prediction. Finally, Section~\ref{sec:conclusion} concludes the paper.
\section{Methodology}
\label{sec:method}
\begin{figure}
\centering
\includegraphics[width=\linewidth]{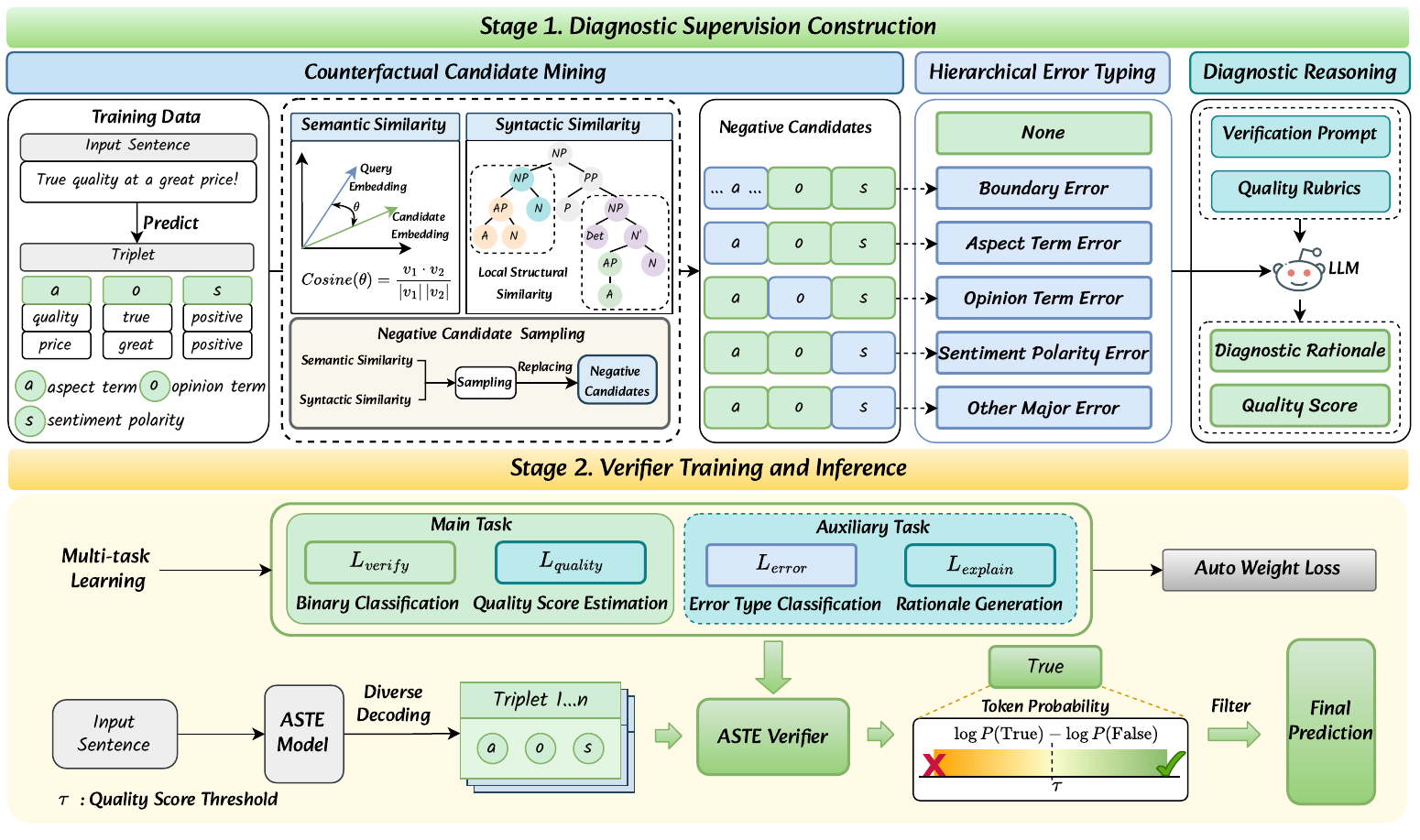}
\caption{Overview of FiVeD for fine-grained ASTE verification. The framework first constructs diagnostic supervision by mining counterfactual candidate triplets, assigning fine-grained error types, and obtaining LLM-generated rationales and quality scores. It then trains a verifier with multi-task learning to rescore and filter candidate triplets produced by a base ASTE model.}
\label{fig:framework}
\end{figure}

Figure~\ref{fig:framework} illustrates the proposed two-stage FiVeD framework for fine-grained ASTE verification. The first stage, \textbf{Diagnostic Supervision Construction}, builds a diagnostic verification dataset from gold ASTE annotations. Specifically, we perform \textbf{Counterfactual Candidate Mining} to synthesize plausible negative triplets, assign each candidate a specific error type through \textbf{Hierarchical Error Typing}, and obtain soft quality scores and diagnostic rationales through \textbf{Diagnostic Reasoning} from an off-the-shelf LLM. The second stage, \textbf{Verifier Training and Inference}, trains an ASTE verifier with multi-task learning. The verifier jointly learns binary verification, quality score estimation, error type classification, and rationale generation. During inference, the trained verifier rescores candidate triplets produced by a base ASTE model and filters them based on a quality score threshold \(\tau\).

\subsection{Task Formulation}

Given an input sentence $x = \{w_1, w_2, \ldots, w_n\}$, the aspect sentiment triplet extraction task aims to extract a set of triplets:
\begin{equation}
\mathcal{Y} = \{y_i\}_{i=1}^{N},
\end{equation}
where each triplet is defined as:
\begin{equation}
y_i = (a_i, o_i, s_i).
\end{equation}
Here, $a_i$ denotes an aspect term, $o_i$ denotes an opinion term, and $s_i \in \{\text{positive}, \text{negative}, \text{neutral}\}$ denotes the sentiment polarity.

Instead of directly accepting all triplets predicted by an ASTE model, we formulate a verification problem. Given a sentence $x$ and a candidate triplet $\hat{y} = (\hat{a}, \hat{o}, \hat{s})$, the ASTE verifier estimates whether $\hat{y}$ is a correct extraction result and assigns a normalized quality score:
\begin{equation}
q_{\theta}(x,\hat{y}) \in [0,1].
\end{equation}

The final prediction set is obtained by filtering candidate triplets according to the quality score estimated by the verifier. Formally, given a candidate set $\hat{\mathcal{Y}}$ generated by a base ASTE model, the final output is:
\begin{equation}
\mathcal{Y}^{*}
=
\{
\hat{y} \in \hat{\mathcal{Y}}
:
q_{\theta}(x,\hat{y}) \ge \tau
\},
\end{equation}
where $\tau$ is a score threshold that controls the precision-recall trade-off.

\subsection{Stage 1: Diagnostic Supervision Construction}

The goal of Stage 1 is to construct diagnostic verification data for training the ASTE verifier. A naive verifier trained only with valid candidates and randomly corrupted invalid candidates may fail to identify realistic ASTE errors. Therefore, we construct negative candidates through \textbf{Counterfactual Candidate Mining}, assign fine-grained diagnostic labels through \textbf{Hierarchical Error Typing}, and obtain soft quality scores and rationales through \textbf{Diagnostic Reasoning} from LLMs.

Each constructed training instance is represented as $(x, \hat{y}, v, q, e, r)$, where $x$ is the input sentence, $\hat{y}$ is a candidate triplet, $v \in \{True, False\}$ is the binary validity label, $q \in [0,1]$ is the estimated quality score, $e$ is the hierarchical error type, and $r$ is the diagnostic rationale.

\subsubsection{Training Data}

We start from gold ASTE training data. For each input sentence $x$, the gold annotations contain one or more triplets $\mathcal{Y} = \{(a_i, o_i, s_i)\}_{i=1}^{N}.$ For example, given the sentence \textit{``True quality at a great price!''}, the gold triplets may include $(\text{quality}, \text{true}, \text{positive})$ and $(\text{price}, \text{great}, \text{positive})$. Gold triplets are treated as positive candidates with a validity label $v=\text{True}$. We then generate negative candidates by applying controlled counterfactual perturbations to these triplets in the following section.

\subsubsection{Counterfactual Candidate Mining}

This module synthesizes negative candidates that are close to gold triplets but contain specific extraction errors. As shown in Figure~\ref{fig:framework}, it first mines plausible distractors from the input sentence, then applies controlled perturbations to gold triplets, and finally filters and subsamples the generated candidates to obtain informative negatives.

Given a sentence \(x\) and a gold triplet \(y=(a,o,s)\), where \(a\), \(o\), and \(s\) denote the aspect term, opinion term, and sentiment polarity, respectively, we generate a counterfactual candidate:
\begin{equation}
\hat{y}=(\hat{a},\hat{o},\hat{s}),
\end{equation}
by altering one or more components of \(y\). The perturbations are designed to approximate common ASTE errors, including aspect term errors, opinion term errors, boundary errors, sentiment polarity errors, cross-triplet mismatches, and compound errors involving multiple incorrect components.

\paragraph{Element-level Candidate Mining.}
We construct plausible element-level candidates by combining syntactic constraints and semantic similarity. For aspect terms, we mainly rely on syntactic plausibility. Specifically, we parse each sentence with a dependency parser and use part-of-speech consistency and token proximity to identify aspect distractors, since aspect errors often arise from confusing the gold aspect with nearby nouns or noun phrases.

For opinion terms, we first use syntactic cues to extract opinion-like candidates, such as adjective- or adverb-headed phrases derived from the dependency parse. We then rank these candidates according to their semantic similarity to the gold opinion term. Given the embedding of the gold opinion span \(u_o\) and the embedding of an opinion candidate \(v_{\hat{o}}\), semantic similarity is computed as:
\begin{equation}
\operatorname{sim}_{sem}(u_o,v_{\hat{o}})
=
\frac{u_o^{\top}v_{\hat{o}}}{\|u_o\|\|v_{\hat{o}}\|}.
\end{equation}
This strategy ensures that opinion distractors are both locally plausible and semantically related to the gold opinion, producing hard negatives that are more informative for verifier training.

\paragraph{Counterfactual Candidate Sampling.}
Starting from each complex gold triplet that co-occurs with at least one other gold triplet in the same sentence, we synthesize counterfactual candidates through controlled edits. Sentences with only one gold triplet are excluded from counterfactual mining because they offer little room for plausible confusion and excluding them helps maintain the positive–negative balance of the verification corpus. As summarized in Algorithm~\ref{alg:candidate-mining}, the edits include aspect term replacement, opinion term replacement, aspect and opinion boundary perturbation, sentiment polarity flipping, compound perturbations, and cross-triplet mispairing. Each edit is associated with an error label under our fine-grained annotation schema.

\begin{algorithm}[t]
\caption{Counterfactual Negative Candidate Mining}
\label{alg:candidate-mining}
\begin{algorithmic}[1]
\REQUIRE Sentence \(x\), gold triplets \(\mathcal{Y}\), sentiment set \(\mathcal{S}\), sampling budget \(K\)
\ENSURE Negative candidate set \(\hat{\mathcal{Y}}^{-}\)

\IF{\(|\mathcal{Y}|=1\)}
\STATE \textit{Single-triplet sentences are treated as simple cases.}
    \RETURN \(\emptyset\)
\ENDIF

\STATE Parse \(x\) with a dependency parser
\STATE Extract aspect replacement candidates \(\mathcal{A}(x)\) using syntactic constraints
\STATE Extract opinion replacement candidates \(\mathcal{O}_{syn}(x)\) using syntactic cues
\STATE Initialize \(\hat{\mathcal{Y}}^{-}\leftarrow \emptyset\)

\FORALL{\((a,o,s)\in\mathcal{Y}\)}
    \STATE Rank \(\mathcal{O}_{syn}(x)\) by semantic similarity to \(o\)
    \STATE Initialize candidate sets \(\mathcal{C}_{a}, \mathcal{C}_{o}, \mathcal{C}_{s}, \mathcal{C}_{bnd}, \mathcal{C}_{other}\leftarrow \emptyset\)

    \STATE Add aspect replacement candidates \((\hat{a},o,s)\) to \(\mathcal{C}_{a}\), where \(\hat{a}\in\mathcal{A}(x)\)
    \STATE Add opinion replacement candidates \((a,\hat{o},s)\) to \(\mathcal{C}_{o}\), where \(\hat{o}\) is top-ranked in \(\mathcal{O}_{syn}(x)\)
    \STATE Add aspect/opinion-boundary variants \((\tilde{a},o,s)\) and \((a,\tilde{o},s)\) to \(\mathcal{C}_{bnd}\)
    \STATE Add sentiment-flipped candidates \((a,o,\hat{s})\) to \(\mathcal{C}_{s}\), where \(\hat{s}\in\mathcal{S}\setminus\{s\}\)
    \STATE Add compound candidates to \(\mathcal{C}_{other}\) by combining at least two edit types

    \FORALL{\((a_j,o_j,s_j)\in\mathcal{Y}\setminus\{(a,o,s)\}\)}
        \STATE Construct cross-triplet candidate \(\hat{y}_{j}=(a,o_j,s)\)
        \IF{\(\hat{y}_{j}\) differs in at least two components from every gold triplet}
            \STATE \(\mathcal{C}_{other}\leftarrow \mathcal{C}_{other}\cup\{\hat{y}_{j}\}\)
        \ENDIF
    \ENDFOR

    \STATE Remove duplicate or gold-matching candidates from each candidate set
    \STATE Subsample up to \(K\) candidates from each of \(\mathcal{C}_{a}, \mathcal{C}_{o}, \mathcal{C}_{s}, \mathcal{C}_{bnd}, \mathcal{C}_{other}\)
    \STATE \(\mathcal{C}\leftarrow \mathcal{C}_{a}\cup \mathcal{C}_{o}\cup \mathcal{C}_{s}\cup \mathcal{C}_{bnd}\cup \mathcal{C}_{other}\)
    \STATE \(\hat{\mathcal{Y}}^{-}\leftarrow \hat{\mathcal{Y}}^{-}\cup \operatorname{Subsample}(\mathcal{C}, K)\)
\ENDFOR

\STATE Deduplicate \(\hat{\mathcal{Y}}^{-}\)
\RETURN \(\hat{\mathcal{Y}}^{-}\)
\end{algorithmic}
\end{algorithm}
Aspect term replacement substitutes the gold aspect with syntactically plausible distractors, while opinion term replacement substitutes the gold opinion with syntactically valid and semantically similar opinion candidates. Boundary errors are generated by shifting, expanding, or shrinking the aspect or opinion span within a small token window, retaining only variants that partially overlap with the original span. Sentiment errors are obtained by replacing the gold sentiment with an alternative polarity label. We further generate compound and structural errors by combining multiple perturbation axes or by pairing aspects and opinions from different gold triplets. Such candidates are retained as \emph{other major} errors only if they differ in at least two components from every gold triplet in the sentence.

\paragraph{Filtering and Subsampling.}
After synthesis, we filter the candidate pool to remove trivial, duplicate, or structurally invalid candidates. Specifically, we discard candidates that exactly match any gold triplet, candidates with empty or malformed spans, and candidates that violate basic linguistic constraints. For compound and cross-triplet candidates, we additionally apply the two-component-difference rule described above.

To control the size of the verification corpus, we apply randomized subsampling with a fixed per-gold budget \(K\). This strategy does not enforce exact class proportions, but it prevents easily enumerable perturbations from dominating the training data and encourages the retention of diverse error mechanisms. Then, for a sentence \(x\) with more than one gold triplet, denoted as
\(\mathcal{Y}(x)=\{y_i\}_{i=1}^{N_x}\) with \(N_x>1\), the module generates up to \(K\) counterfactual negative candidates for each gold triplet:
\begin{equation}
\hat{\mathcal{Y}}^{-}_i(x)
=
\{\hat{y}^{-}_{i,k}\}_{k=1}^{K_i},
\qquad K_i\leq K.
\end{equation}
The sentence-level negative candidate set is then obtained by aggregating candidates over all gold triplets:
\begin{equation}
\hat{\mathcal{Y}}^{-}(x)
=
\bigcup_{i=1}^{N_x}
\hat{\mathcal{Y}}^{-}_i(x).
\end{equation}
Together with the original gold triplets \(\mathcal{Y}(x)\) as positive candidates, the mined negatives form counterfactual instances, which are aggregated over all training sentences to build the final verification corpus.

\subsubsection{Hierarchical Error Typing}

After constructing candidate triplets, we assign each candidate a diagnostic error type through Hierarchical Error Typing. 
The goal is to enrich the verification signal beyond a binary correctness label by explicitly characterizing the nature and severity of the candidate error.

Given the sentence \(x\), the gold triplet set \(\mathcal{Y}(x)\), and the candidate triplet \(\hat{y}\), we assign an error label $e \in \mathcal{E}$, where the error taxonomy is defined as:
\begin{equation}
\begin{aligned}
\mathcal{E}
=
\{&
\textsc{None},
\textsc{Boundary Error},
\textsc{Aspect Term Error},\\
&
\textsc{Opinion Term Error},
\textsc{Sentiment Polarity Error},
\textsc{Other Major Error}
\}.
\end{aligned}
\end{equation}

\begin{table}[t]
\centering
\small
\begin{tabular}{p{0.22\linewidth} p{0.74\linewidth}}
\toprule
\textbf{Error Type} & \textbf{Definition} \\
\midrule
\textsc{None} 
& The prediction is completely correct: the aspect term, opinion term, and sentiment polarity match the gold triplet. \\
\midrule

\textsc{Boundary Error} 
& The predicted span has token-level overlap with the corresponding gold span, but its boundary is not exact, e.g., a subset, superset, or shifted span. \\
\midrule
\textsc{Aspect Term Error} 
& The predicted aspect term is incorrect and has no token-level overlap with the gold aspect term. \\
\midrule
\textsc{Opinion Term Error} 
& The predicted opinion term is incorrect and has no token-level overlap with the gold opinion term. \\
\midrule
\textsc{Sentiment Polarity Error} 
& The aspect and opinion terms are correct, but the sentiment polarity is incorrect. \\
\midrule
\textsc{Other Major Error} 
& The prediction contains other major structural errors, including incorrect aspect--opinion pairing, fabricated terms, cross-triplet mismatching, or multiple compounded errors. \\
\bottomrule
\end{tabular}
\caption{Diagnostic error taxonomy for candidate triplet verification.}
\label{tab:error_taxonomy}
\end{table}

The error taxonomy is organized hierarchically as an operational schema, summarized in Table~\ref{tab:error_taxonomy} and illustrated with examples provided in Figure~\ref{fig:example}. This is motivated by common ABSA error patterns~\cite{lai2025E4L,rvisa}, where boundary errors and single-element errors are often dominant. It first distinguishes fully correct candidates from erroneous ones, and then categorizes erroneous candidates by their primary error source, including span boundary mismatch, aspect term mismatch, opinion term mismatch, sentiment polarity mismatch, and other structural failures. We group cross-triplet mismatches and compounded failures into \textsc{Other Major Error}, as they usually involve multiple interacting components and cannot be reliably attributed to a single element-level source. This design keeps the taxonomy compact while preserving the distinction among minor boundary deviations, single-element errors, and severe structural failures. 


\subsubsection{Diagnostic Reasoning}

The Diagnostic Reasoning module produces estimated quality scores and natural-language rationales for candidate triplets. 
As shown in Figure~\ref{fig:framework}, this module combines deterministic error typing with LLM-based diagnostic assessment. 
The deterministic component first assigns an error type \(e\), which constrains the major error family of the candidate. 
Conditioned on this error type, an LLM reasoner is then prompted to produce a graded quality score and a concise diagnostic rationale.

Given a sentence \(x\), the gold triplet set \(\mathcal{Y}(x)\), a candidate triplet \(\hat{y}_i\), its binary validity label \(v_i\), and its error type \(e_i\), the LLM reasoner produces:
\begin{equation}
(q_i, r_i) = f_{\mathrm{LLM}}(x, \mathcal{Y}(x), \hat{y}_i, v_i, e_i),
\end{equation}
where \(q_i \in [0,1]\) denotes the quality score and \(r_i\) denotes the natural-language diagnostic rationale.

We adopt a 4-level scoring rubric summarized in Table~\ref{tab:quality_rubric}, which is conceptually analogous to the graded relevance convention in information retrieval evaluation \cite{Kalervo_IR}. Following this convention, the cutoffs are used as ordinal anchors to reflect the degree of validity of a candidate triplet rather than calibrated probabilities. 
Specifically, \(q=1.0\) corresponds to a completely correct prediction, higher scores indicate minor form-level deviations, intermediate scores indicate partial semantic alignment with single-element error, and lower scores indicate severe semantic or structural failures. 
The specific numerical boundaries provide deterministic intervals within which the LLM assigns fine-grained scores reflecting error severity, while the ordering and partition structure, rather than the specific values, drive verifier learning.

\begin{table}[t]
\centering
\small
\begin{tabular}{p{0.12\linewidth} p{0.80\linewidth}}
\toprule
\textbf{Score Range} & \textbf{Interpretation} \\
\midrule
\(q=1.0\)
& Completely correct prediction. The meaning and all required triplet elements match the gold annotation. \\
\midrule
\(0.7 \le q < 1.0\) 
& Minor errors. The prediction is essentially correct, but contains some minor formal deviations, such as a boundary mismatch with token overlap. \\
\midrule
\(0.3 \le q < 0.7\)
& Moderate errors. Exactly one key element is incorrect, but the prediction remains partially aligned with the gold triplet. \\
\midrule
\(0.0 \le q < 0.3\)
& Major errors. The prediction is semantically unsupported or substantially different, such as incorrect aspect--opinion pairing, fabricated content, or multiple compounded errors. \\
\bottomrule
\end{tabular}
\caption{Quality score rubric used for LLM estimation, where $q$ denotes the quality score. The score ranges serve as ordinal anchors reflecting the degree of validity of a candidate triplet rather than calibrated probabilities.}
\label{tab:quality_rubric}
\end{table}

In addition to the quality score, the LLM is required to output an explanatory diagnostic rationale. 
The rationale explains why the candidate is correct or incorrect by comparing the candidate triplet against the gold triplet set in the context of the original sentence. 
It focuses on the linguistic and semantic relationship between the prediction and the gold annotation, without explicitly referring to the numerical score. 
This rationale serves as an auxiliary supervision signal, encouraging the verifier to associate triplet predictions with explicit linguistic and semantic evidence.

To ensure output consistency, we enforce a structured JSON format containing two fields: \texttt{quality\_score} and \texttt{rationale}. 
The former is parsed as the quality score \(q\), while the latter is used as the diagnostic rationale \(r\). 
The final output of Stage~1 is a multi-objective verification corpus:
\begin{equation}
\mathcal{D}_{\mathrm{ver}}
=
\{
(x_i,\mathcal{Y}(x_i), \hat{y}_i, v_i, q_i, e_i, r_i)
\}_{i=1}^{|\mathcal{D}_{\mathrm{ver}}|},
\end{equation}
where \(x_i\) is the sentence, \(\mathcal{Y}(x_i)\) is the gold triplet set, \(\hat{y}_i\) is a candidate triplet, \(v_i\) is the binary verification label, \(q_i\) is the quality score, \(e_i\) is the diagnostic error type, and \(r_i\) is the diagnostic rationale. 
This corpus is subsequently used to train the ASTE verifier in Stage~2.

\subsection{Stage 2: Verifier Training and Inference}

Stage~2 trains an ASTE verifier using the verification corpus constructed in Stage~1. 
We instantiate the verifier as an encoder-decoder model with task-specific prompts. 
This text-to-text formulation is particularly suitable for verification, as the encoder supports semantic understanding, while the decoder provides flexible conditional generation for triplet validity, error types, and diagnostic rationales.

Task-specific prefixes are prepended to the serialized input to specify the target behavior. 
The verifier is trained under a multi-task learning framework, with \textbf{Validity Verification} and \textbf{Quality Score Estimation} as primary tasks, and \textbf{Error Type Prediction} (ETC) and \textbf{Rationale Generation} (RG) as auxiliary tasks. 
All tasks share the same sequence-to-sequence backbone, while their relative contributions are automatically learned during training.

\subsubsection{Validity Verification}

For validity verification, the verifier determines whether a candidate triplet is valid with respect to the input sentence. 
Following the sequence-to-sequence formulation, the validity label is verbalized as
\begin{equation}
v \in \{\mathrm{True}, \mathrm{False}\}.
\end{equation}
The model is trained to generate \(v\) using the standard teacher-forced negative log-likelihood loss:
\begin{equation}
\mathcal{L}_{\mathrm{verify}}
=
-
\sum_{j=1}^{|v|}
\log p_{\theta}
\left(
v_j \mid v_{<j}, x, \hat{y}
\right).
\end{equation}
where \(v_j\) is the \(j\)-th token of the target decision. 
This objective preserves the generative nature of the verifier and avoids introducing a separate binary classification head.

\subsubsection{Quality Score Estimation}

Quality score estimation provides a continuous target \(q \in [0,1]\) for candidate-level correctness. 
Instead of adding a regression head, we derive a verification logit directly from the sequence-to-sequence model's generative probabilities over the verbalized decisions:
\begin{equation}
\label{eq:prob}
z_{\theta}(x,\hat{y})
=
\log P_{\theta}(v=\mathrm{True} \mid x,\hat{y})
-
\log P_{\theta}(vv=\mathrm{False} \mid x,\hat{y}),
\end{equation}
where \(P_{\theta}(v=\mathrm{True}\mid x,\hat{y})\) and \(P_{\theta}(v=\mathrm{False}\mid x,\hat{y})\) denote the autoregressive sequence probabilities of generating the verbalized validity labels ``True'' and ``False'', respectively.

This relative log-probability measures the model's preference for accepting the candidate over rejecting it. 
We then optimize \(z_\theta\) with a soft-target binary cross-entropy loss:
\begin{equation}
\mathcal{L}_{\mathrm{quality}}
=
-\mathbb{E}_{(x,\hat{y},q)}\Big[
q \log \sigma\!\left(z_{\theta}(x,\hat{y})\right)
+
(1-q)\log\!\left(1-\sigma\!\left(z_{\theta}(x,\hat{y})\right)\right)
\Big].
\end{equation}
Compared with a mean-squared regression loss, this formulation is better aligned with the verification setting by treating \(q\) as a soft target and directly operating on the relative preference between \(\mathrm{True}\) and \(\mathrm{False}\) induced by sequence-to-sequence generative probabilities.
In this way, \(\mathcal{L}_{\mathrm{quality}}\) complements the hard label supervision of \(\mathcal{L}_{\mathrm{verify}}\), providing graded supervision over candidate quality.

\subsubsection{Error Type Classification}

Error type classification is an auxiliary prediction task. 
Given the diagnostic error label \(e\) from Stage~1, the verifier is trained to generate its textual form:
\begin{equation}
\mathcal{L}_{\mathrm{error}}
=
-
\sum_{j=1}^{|e|}
\log p_{\theta}
\left(
e_j \mid e_{<j}, x, \hat{y}
\right).
\end{equation}

This auxiliary objective encourages the verifier to distinguish between different failure modes, rather than simply determining triplet validity.

\subsubsection{Rationale Generation}

Rationale generation is another auxiliary task for explaining the reason behind the validity judgment. 
Given the diagnostic rationale \(r\), the verifier is trained with:
\begin{equation}
\mathcal{L}_{\mathrm{explain}}
=
-
\sum_{j=1}^{|r|}
\log p_{\theta}
\left(
r_j \mid r_{<j}, x, \hat{y}, e
\right).
\end{equation}

The rationale serves as an auxiliary training signal that encourages the verifier to ground its decision in explicit linguistic and semantic discrepancies between the candidate and the reference annotation. 

\subsubsection{Automatic Loss Weighting}

The four objectives have different scales, learning dynamics, and levels of supervision granularity. 
We therefore combine them using automatic task weighting rather than manually assigning fixed coefficients. 
Let the task set
\begin{equation}
\mathcal{T}
=
\{
\mathrm{verify},
\mathrm{quality},
\mathrm{error},
\mathrm{explain}
\}.
\end{equation}
The overall objective is:
\begin{equation}
\mathcal{L} =
\sum_{t \in \mathcal{T}}
\left(
\frac{1}{2\sigma_t^2}
\mathcal{L}_t
+
ln({\sigma_t}^2 + 1)
\right),
\end{equation}
where \(\sigma_t\) is a learnable uncertainty parameter for task $t$. 

In this formulation, the effective weight of task \(t\) is controlled by \(\frac{1}{2\sigma_t^2}\), so tasks with larger estimated uncertainty contribute less to the overall objective, while tasks with lower uncertainty are assigned larger weights. The logarithmic regularization term \(\ln(\sigma_t^2+1)\) helps avoid negative values and keeps the objective numerically stable, following \cite{multi2/abs-1805-06334}. 
As a result, the relative contributions of validity verification, quality score estimation, error type classification, and rationale generation are learned jointly with the verifier parameters, avoiding manual loss-weight tuning.

\subsection{Inference with the ASTE Verifier}

During inference, the framework coordinates the base ASTE model with the trained verifier. Given an input sentence \(x\), the base ASTE model first generates a candidate multiset through one or multiple decoding runs:
\begin{equation}
\hat{\mathcal{Y}}^{\mathrm{multi}}
=
\{
\hat{y}_1,\hat{y}_2,\ldots,\hat{y}_M
\}.
\end{equation}

To improve coverage, diverse decoding may be used to produce multiple candidate triplets. Since the same triplet may be generated multiple times across different decoding paths, we first estimate the empirical support of each unique candidate. For each candidate \(\hat{y}\), its support count is defined as:
\begin{equation}
c(\hat{y})
=
\sum_{m=1}^{M}
\mathbb{I}
\left[
\hat{y}_m = \hat{y}
\right],
\end{equation}
where \(\mathbb{I}[\cdot]\) denotes the indicator function.

We then apply \textbf{Candidate Support Filtering} (CSF) to remove low-frequency candidates:
\begin{equation}
\hat{\mathcal{Y}}_{c_{\min}}
=
\left\{
\hat{y} \in \mathrm{Unique}
\left(
\hat{\mathcal{Y}}^{\mathrm{multi}}
\right)
:
c(\hat{y}) \ge c_{\min}
\right\},
\end{equation}
where \(c_{\min}\) is the minimum support threshold. This step filters out candidates that are only sporadically generated, thereby reducing noisy or unstable predictions before verification.

Each remaining candidate \(\hat{y}\in\hat{\mathcal{Y}}_{c_{\min}}\) is then scored by the verifier using the relative log-probability \(z_{\theta}(x,\hat{y})\)  between the two verbalized decisions, defined in the same way as Equation \ref{eq:prob}.
The normalized verification score is:
\begin{equation}
\hat{q}_{\theta}(x,\hat{y})
=
\sigma
\left(
z_{\theta}(x,\hat{y})
\right).
\end{equation}

The final prediction set is obtained by applying the verification threshold:
\begin{equation}
\mathcal{Y}^{*}
=
\{
\hat{y} \in \hat{\mathcal{Y}}_{c_{\min}}
:
\hat{q}_{\theta}(x,\hat{y}) \ge \tau
\}.
\end{equation}

The minimum support threshold \(c_{\min}\) and the verification threshold \(\tau\) control different aspects of inference. The support threshold \(c_{\min}\) filters candidates according to their generation frequency across decoding runs, removing low-support candidates before verification. In contrast, the verification threshold \(\tau\) controls the trade-off between precision and recall. A higher \(c_{\min}\) or \(\tau\) yields more conservative predictions and tends to improve precision, whereas lower thresholds retain more candidates and tend to improve recall.

\section{Experiments}
\label{experiment}

\subsection{Setup}

\paragraph{\textbf{Datasets and Evaluation Metrics.}}
We conduct experiments on four standard benchmark datasets widely 
used for ASTE evaluation, including Rest14, Rest15, Rest16, and Lap14, which are derived from SemEval-2014, SemEval-2015, and SemEval-2016 \cite{Pontiki_semeval_2014, pontiki-etal-2015-semeval, pontiki-etal-2016-semeval}. Following prior studies \cite{Peng_Xu_Bing_Huang_Lu_Si_2020, xu-etal-2020-position}, we use the same data splits to ensure fair comparison. Dataset statistics are reported in Table~\ref{tab:dataset_statistics}. We evaluate model performance using precision (P), recall (R), and F1 score.

\begin{table*}[t]
\centering
\begin{minipage}[t]{0.45\textwidth}
\centering
\captionof{table}{Statistics of ASTE datasets. \#S and \#T denote the number of sentences and triplets, respectively.}
\label{tab:dataset_statistics}
\resizebox{0.95\linewidth}{!}{
\begin{tabular}{lcccccc}
\toprule
\multirow{2}{*}{Dataset} 
& \multicolumn{2}{c}{Train} 
& \multicolumn{2}{c}{Dev} 
& \multicolumn{2}{c}{Test} \\
\cmidrule(lr){2-3}
\cmidrule(lr){4-5}
\cmidrule(lr){6-7}
& \#S & \#T & \#S & \#T & \#S & \#T \\
\midrule
Rest14 & 1266 & 2338 & 310 & 577 & 492 & 994 \\
Rest15 & 605  & 1013 & 148 & 249 & 322 & 485 \\
Rest16 & 857  & 1394 & 210 & 339 & 326 & 514 \\
Lap14  & 906  & 1460 & 219 & 346 & 328 & 543 \\
\bottomrule
\end{tabular}
}
\end{minipage}
\hfill
\begin{minipage}[t]{0.52\textwidth}
\centering
\captionof{table}{Error type distribution in the constructed verification corpus. The ``None'' category denotes positive examples, while all error categories jointly constitute counterfactual negative examples.}
\label{tab:error_type_distribution}
\resizebox{\linewidth}{!}{
\begin{tabular}{lccccc}
\toprule
Error type & Rest14 & Rest15 & Rest16 & Lap14 & Total \\
\midrule
None & 2338 & 1013 & 1394 & 1460 & 6205 (42\%) \\
\midrule
Boundary Error & 733 & 184 & 247 & 309 & \multirow{5}{*}{8426 (58\%)} \\
Aspect Term Error & 333 & 162 & 196 & 221 &  \\
Opinion Term Error & 667 & 237 & 353 & 386 &  \\
Sentiment Polarity Error & 829 & 402 & 501 & 451 &  \\
Other Major Error & 904 & 365 & 483 & 463 &  \\
\bottomrule
\end{tabular}
}
\end{minipage}
\end{table*}

\paragraph{\textbf{Baselines.}}
We compare our method with representative ASTE methods from five categories: sequence tagging, MRC-based methods, span-based methods, table tagging methods, and generative methods. For non-generative baselines, we report published results from the corresponding papers when available. For generative baselines, we apply our method to three representative models, namely Paraphrase \cite{emnlp/ZhangD0YBL21}, GAS \cite{acl/Zhang0DBL20}, and MvP \cite{acl/MvP}, to evaluate its plug-and-play verification capability.

\textbf{a) Sequence tagging methods.}
\begin{itemize}
    \item Peng-two-stage \cite{Peng_Xu_Bing_Huang_Lu_Si_2020} decomposes ASTE into aspect extraction, aspect term sentiment classification, and opinion term extraction under a two-stage framework. 
    \item Span-BART \cite{yan-etal-2021-unified} formulates ABSA subtasks as a unified sequence-to-sequence generation problem with BART.
    \item TAGS \cite{xianlong-etal-2023-tagging} employs an additional sequence tagging task to enhance the generation task under encoder and decoder supervision.
\end{itemize}

\textbf{b) MRC-based methods.}
\begin{itemize}
    \item Dual-MRC \cite{Dual-MRC21} casts ASTE as two machine reading comprehension (MRC) problems and jointly trains parameter-shared BERT-MRC models.
    \item COM-MRC \cite{zhai-etal-2022-com} enhances the MRC formulation with contextual augmentation.
    \item Triple-MRC \cite{Zou2024AMS} introduces a multi-task cascaded MRC framework for ASTE.
\end{itemize}

\textbf{c) Span-based methods.}
\begin{itemize}
    \item Span-ASTE \cite{xu-etal-2021-learning} learns span-level representations and uses dual-channel span pruning with auxiliary aspect and opinion supervision.
    \item D2E2S \cite{zhao-etal-2024-dual} combines a BERT-based semantic encoder with an enhanced LSTM channel to capture syntactic information.
\end{itemize}

\textbf{d) Table tagging methods.}
\begin{itemize}
    \item EMC-GCN \cite{chen-etal-2022-enhanced} models word pair relations with graph convolution networks and incorporates diverse linguistic features.
    \item SimSTAR \cite{SIMSTAR} decouples sentiment extraction from term extraction with a segment tagging scheme and dual extractors.
    \item MiniConGTS \cite{sun-etal-2024-minicongts} improves grid-tagging-based ASTE with a compact and effective table representation.
\end{itemize}

\textbf{e) Generative methods.}
\begin{itemize}
    \item GPT-3.5-turbo and GPT-4o are evaluated under zero-shot, few-shot, Chain-of-Thought (CoT), and CoT with few-shot prompting settings.
    \item LEGO \cite{coling/GaoFLLLLBY22} unifies multiple ABSA tasks through task prompts composed of multiple elements.
    \item Paraphrase \cite{emnlp/ZhangD0YBL21} reformulates structured sentiment extraction as natural language generation with semantic guidance.
    \item GAS \cite{acl/Zhang0DBL20} uses a T5-based generative architecture with annotation-style and extraction-style training paradigms.
    \item MvP \cite{acl/MvP} introduces element-order perturbation to improve multi-view prompting for generative ASTE models.
\end{itemize}

\paragraph{\textbf{Models and Hyperparameters.}}
To construct the verification corpus, we use off-the-shelf LLMs to provide diagnostic rationales and quality score estimates. Specifically, we experiment with Qwen2.5-72B-Instruct and DeepSeek-Reasoner as diagnostic reasoners. The sampling temperature is set to \(T=0.7\) following \cite{nips/refine} to encourage diverse yet reliable diagnostic rationales under gold label guidance. Table~\ref{tab:error_type_distribution} summarizes the error type distribution in the constructed verification corpus.
The ``None'' category denotes positive verification instances from source data, while the remaining categories correspond to counterfactual negative instances whose counts are kept relatively balanced with the positive instances.
The verifier is built upon FLAN-T5-large \cite{flant5} in an encoder-decoder architecture. We train the verifier with a batch size of \(64\) and a learning rate of \(1 \times 10^{-5}\). AdamW is used as the optimizer, together with a linear learning-rate decay scheduler. The maximum number of training epochs is set to \(20\), and early stopping is applied with a patience of \(10\) epochs based on the validation F1 score. Since we adopt automatic weighted loss (AWL), the weights of different auxiliary objectives are jointly optimized with the model parameters rather than manually tuned.

We consider two training settings for the verifier. In the dataset-specific setting, a separate verifier is trained for each dataset. In the unified setting, a single verifier is trained on the combined training data from all datasets, reducing training costs and improving cross-domain utilization. During inference, the minimum support count threshold \(c_{\min}\) and the verifier quality score threshold \(\tau\) are selected on the validation set according to the best F1 score.

\begin{table*}[h]
  \centering
  \resizebox{\textwidth}{!}{%
  \begin{tabular}{lccccccccccccc}
    \toprule
    \cmidrule(lr){2-13}
    \multirow{2}{*}{\textbf{Methods}}
      & \multicolumn{3}{c}{\textbf{Rest14}}
      & \multicolumn{3}{c}{\textbf{Rest15}}
      & \multicolumn{3}{c}{\textbf{Rest16}}
      & \multicolumn{3}{c}{\textbf{Lap14}} & \multirow{2}{*}{\makecell{\textbf{Avg}\\\textbf{F1}}}\\
    \cmidrule(lr){2-4}\cmidrule(lr){5-7}\cmidrule(lr){8-10}\cmidrule(lr){11-13}
      & P & R & F1 & P & R & F1 & P & R & F1 & P & R & F1 & \\
    \midrule
    \multicolumn{14}{l}{\textit{- Sequence tagging}} \\
    Peng-two-stage \cite{Peng_Xu_Bing_Huang_Lu_Si_2020}
      & 43.24 & 63.66 & 51.46 & 48.07 & 57.51 & 52.32 & 46.96 & 64.24 & 54.21 & 37.38 & 50.38 & 42.87 & 50.22 \\
    Span-BART \cite{yan-etal-2021-unified}
      & 65.52 & 64.99 & 65.25 & 59.14 & 59.38 & 59.26 & 66.60 & 68.68 & 67.62 & 61.41 & 56.19 & 58.69 & 62.71 \\
    TAGS \cite{xianlong-etal-2023-tagging}
      & 74.92 & 73.81 & 74.36 & 69.55 & 65.25 & \underline{67.33} & 75.40 & 72.48 & 74.17 & 64.69 & 61.89 & 63.26 & 69.78 \\
    \midrule
    \multicolumn{14}{l}{\textit{- MRC-based}} \\
    Dual-MRC \cite{Dual-MRC21} &71.55& 69.14& 70.32& 63.78& 51.87& 57.21& 68.60 &66.24& 67.40 &57.39 &53.88 &55.58& 62.63\\
    COM-MRC  \cite{zhai-etal-2022-com}
      & 75.46 & 68.91 & 72.01 & 68.35 & 61.24 & 64.53 & 71.55 & 71.59 & 71.57 & 62.35 & 58.16 & 60.17 & 67.07 \\
    Triple-MRC \cite{Zou2024AMS}
      & / & / & 72.45 & / & / & 62.86 & / & / & 68.65 & / & / & 60.72 & 66.17 \\
    \midrule
    \multicolumn{14}{l}{\textit{- Span-based}} \\
    Span-ASTE \cite{xu-etal-2021-learning}
      & 72.89 & 70.89 & 71.85 & 62.18 & 64.45 & 63.27 & 69.45 & 71.17 & 70.26 & 63.44 & 55.84 & 59.38 & 66.19 \\
    D2E2S \cite{zhao-etal-2024-dual}
      & 75.92 & 74.36 & 75.13 & 70.09 & 62.11 & 65.86 & 77.97 & 71.77 & 74.74 & 67.38 & 60.31 & 63.65 & \underline{69.84} \\
    \midrule
    \multicolumn{14}{l}{\textit{- Table tagging}} \\
    EMC-GCN \cite{chen-etal-2022-enhanced}  &71.21 & 72.39 & 71.78 & 61.70 & 56.26 & 58.81  &61.54 & 62.47 & 61.93  &65.62 & 71.30 & \underline{68.33} & 65.21\\
    SimSTAR \cite{SIMSTAR}
      & 76.23 & 71.63 & 73.86 & 71.71 & 59.59 & 65.09 & 72.07 & 74.12 & 73.06 & 66.46 & 58.23 & 62.07 & 68.52 \\
    MiniConGTS \cite{sun-etal-2024-minicongts}
      & 76.10 & 75.08 & \underline{75.59} & 66.50 & 63.86 & 65.15 & 75.52 & 74.14 & \underline{74.83} & 66.82 & 60.68 & 63.61 & 69.80 \\
    \midrule
    \multicolumn{14}{l}{\textit{- Generative}} \\
    GPT-3.5-turbo $\dagger$ (zero-shot) &44.88 &55.13& 49.48& 36.02& 53.40& 43.02 &39.92 &57.78 &47.22& 30.04& 41.04& 34.69& 43.60\\
    \quad - few-shot& 51.51& 65.19 &57.55& 43.34& 63.09& 51.39 &51.12 &71.01 &59.45& 39.79& 50.09& 44.35& 53.19\\
    \quad - CoT &48.47& 59.05& 53.24 &39.51 &56.70& 46.57& 44.03& 63.81 &52.10& 30.48& 40.30 &34.71 &46.66\\
    \quad - CoT + few-shot &49.41 &59.15 &53.85 & 39.02& 56.08 &46.02 &46.49 &66.93 &54.86&33.78 &42.33& 37.57&48.08\\
    GPT-4o$\dagger$ (zero-shot) &32.99 &38.13 &35.37&  27.85& 37.73 &32.05 &32.17 &43.00& 36.80 &17.81 &22.55 &19.90&31.03\\
    \quad - few-shot &54.11 &66.20& 59.55 & 45.57& 60.41& 51.95& 52.90& 71.01& 60.63 &38.23 &48.61 &42.80&53.73\\
    \quad - CoT &41.21 &53.32 &46.49 &33.07 &50.93 &40.10 &39.14 &58.17 &46.79 &26.98& 37.71 &31.46& 41.21\\
    \quad - CoT + few-shot &46.81& 59.86& 52.54 & 35.08& 53.81 &42.47 &41.53 &61.09 &49.45& 29.71 &40.85& 34.40& 44.72\\
    LEGO \cite{coling/GaoFLLLLBY22}
      & / & / & 73.70 & / & / & 64.40 & / & / & 69.90 & / & / & 62.20 & 67.55 \\
    Paraphrase \cite{emnlp/ZhangD0YBL21}
      & 71.03 & 66.98 & 70.12 & 65.48 & 57.23 & 61.85 & 68.97 & 68.09 & 68.85 & 64.14 & 56.06 & 60.41 & 65.31 \\
    \quad +FiVeD (Qwen)
      & 72.87 & 72.07 & 72.45 & 64.11 & 66.64 & 65.31 & 74.20 & 74.67 & 74.43 & 65.93 & 58.53 & 61.98 & 68.54 \\
    \quad +FiVeD (DeepSeek-Reasoner)
      & 72.20 & 72.80 & 72.48 & 65.11 & 65.65 & 65.36 & 73.65 & 75.29 & 74.45 & 66.87 & 58.16 & 62.21 & 68.63 \\
    \rowcolor{green!12}
    \quad +FiVeD-unified (Qwen)
      & 72.55 & 72.96 & 72.74 & 64.05 & 68.29 & 66.06 & 73.66 & 75.80 & 74.70 & 66.47 & 57.68 & 61.75 & \textbf{68.81} \\
      \rowcolor{green!6}
    \quad $\uparrow$ Improvement
      &  &  & \textcolor{green!45!black}{+2.62}
      &  &  & \textcolor{green!45!black}{+4.21}
      &  &  & \textcolor{green!45!black}{+5.85}
      &  &  & \textcolor{green!45!black}{+1.34}
      & \textcolor{green!45!black}{+3.50} \\
    \quad +FiVeD-unified (DeepSeek-Reasoner)
      & 72.83 & 72.82 & 72.82 & 64.37 & 67.46 & 65.82 & 74.17 & 74.79 & 74.47 & 67.33 & 57.53 & 62.04 & 68.79 \\
    \cmidrule{2-14}
    GAS \cite{acl/Zhang0DBL20}
      & 65.13 & 72.94 & 68.69 & 57.48 & 62.97 & 60.05 & 65.41 & 68.95 & 67.13 & 59.89 & 56.43 & 58.03 & 63.48 \\
    \quad +FiVeD (Qwen)
      & 66.99 & 76.52 & 71.43 & 57.56 & 69.77 & 63.05 & 68.40 & 72.34 & 70.28 & 59.36 & 62.73 & 60.94 & 66.43 \\
    \quad +FiVeD (DeepSeek-Reasoner)
      & 68.36 & 75.86 & 71.91 & 58.15 & 68.91 & 63.01 & 67.42 & 73.27 & 70.13 & 57.99 & 64.09 & 60.86 & 66.48 \\
    \quad +FiVeD-unified (Qwen)
      & 66.78 & 76.68 & 71.38 & 60.98 & 69.98 & 65.13 & 66.23 & 74.47 & 70.01 & 58.82 & 63.35 & 60.88 & 66.85 \\
    \rowcolor{green!12}
    \quad +FiVeD-unified (DeepSeek-Reasoner)
      & 66.41 & 77.23 & 71.40 & 59.74 & 72.37 & 65.39 & 65.17 & 76.26 & 70.25 & 58.33 & 64.16 & 60.99 & \textbf{67.01} \\
      \rowcolor{green!6}
      \quad $\uparrow$ Improvement
      &  &  & \textcolor{green!45!black}{+2.71}
      &  &  & \textcolor{green!45!black}{+5.34}
      &  &  & \textcolor{green!45!black}{+3.12}
      &  &  & \textcolor{green!45!black}{+2.96}
      & \textcolor{green!45!black}{+3.53} \\
    \cmidrule{2-14}
    MvP \cite{acl/MvP}
      & 74.25 & 73.12 & 73.68 & 64.50 & 65.48 & 64.99 & 72.14 & 73.93 & 73.02 & 65.86 & 59.89 & 62.73 & 68.61 \\
    \quad +FiVeD (Qwen)
      & 73.65 & 75.59 & 74.60 & 66.72 & 67.26 & 66.96 & 71.96 & 76.30 & 74.07 & 68.58 & 59.52 & 63.73 & 69.84 \\
    \quad +FiVeD (DeepSeek-Reasoner)
      & 74.95 & 75.01 & 74.94 & 66.59 & 67.26 & 66.87 & 71.80 & 76.46 & 74.05 & 68.09 & 60.14 & 63.86 & 69.93 \\
    \quad +FiVeD-unified (Qwen)
      & 73.36 & 76.04 & 74.67 & 66.49 & 69.28 & 67.82 & 71.61 & 76.93 & 74.17 & 66.76 & 60.78 & 63.60 & 70.07 \\
    \rowcolor{green!12}
    \quad +FiVeD-unified (DeepSeek-Reasoner)
      & 74.15 & 74.79 & 74.46 & 66.95 & 69.24 & 68.03 & 71.66 & 77.12 & 74.29 & 66.42 & 62.51 & 64.38 & \textbf{70.29} \\
      \rowcolor{green!6}
    \quad $\uparrow$ Improvement
      &  &  & \textcolor{green!45!black}{+0.78}
      &  &  & \textcolor{green!45!black}{+3.04}
      &  &  & \textcolor{green!45!black}{+1.27}
      &  &  & \textcolor{green!45!black}{+1.65}
      & \textcolor{green!45!black}{+1.68} \\
    \bottomrule
  \end{tabular}
  }
  \caption{Main results on four ASTE benchmark datasets. Results marked with \(\dagger\) are obtained from \cite{sun-etal-2024-minicongts}. Bold scores highlight the best result enhanced by our method within each generative backbone group, averaged over five runs. The improvement rows report absolute F1 gains over the corresponding base generative model. Underlined F1 scores indicate the best results among previously reported baselines.}
  \label{tab:aste-merged}
\end{table*}

\subsection{Main Results}

Table~\ref{tab:aste-merged} reports the main results on the four ASTE benchmarks. Overall, our method consistently improves all three generative baselines across datasets, demonstrating its effectiveness as a plug-and-play verification module. Compared with the original Paraphrase model, our method achieves average F1 improvements of up to \(+3.50\), with particularly large gains on Rest15 and Rest16. For GAS, our method brings an average F1 improvement of \(+3.53\), improving performance on all datasets and yielding the largest gain on Rest15. For the stronger MvP baseline, our method still improves the average F1 score by \(+1.68\), indicating that the proposed verifier remains beneficial even when the base generator is already competitive.

The improvements mainly come from better recall while maintaining comparable precision. This suggests that our method can recover valid triplets that are missed by single-pass generation, while the verifier prevents excessive false positives through quality score filtering. The effect is especially evident for GAS and Paraphrase, whose original outputs have relatively lower recall or less stable structured generation. By aggregating candidates from multiple decoding runs and verifying them with diagnostic signals, our method produces more reliable final predictions.

We also compare dataset-specific and unified verifier training. The unified setting achieves a competitive or better average F1 in most cases, despite using only a single verifier across all datasets. This indicates that the verifier learns transferable error patterns across domains, such as boundary mismatch, aspect or opinion term errors, and sentiment polarity errors, preserving generalization ability.

Finally, compared with previous non-generative methods, generative models enhanced by our method achieve competitive overall performance. In particular, our method improves MvP to an average F1 of \(70.29\), outperforming the original MvP and surpassing many strong baselines. This confirms the advantage of introducing explicit post-hoc verification into ASTE tasks.

\begin{table*}[h]
\centering
\small
\setlength{\tabcolsep}{3.5pt}
\begin{tabular}{lccccccccccccc}
\toprule
\multirow{2}{*}{Model} 
& \multicolumn{3}{c}{Rest14} 
& \multicolumn{3}{c}{Rest15} 
& \multicolumn{3}{c}{Rest16} 
& \multicolumn{3}{c}{Lap14}
& Avg.\\ 
\cmidrule(lr){2-4}
\cmidrule(lr){5-7}
\cmidrule(lr){8-10}
\cmidrule(lr){11-13}
& P & R & F1 & P & R & F1 & P & R & F1 & P & R & F1 & F1 \\
\midrule

MvP \cite{acl/MvP} & 74.25 & 73.12 & 73.68 & 64.50 & 65.48 & 64.99 & 72.14 & 73.93 & 73.02 & 65.86 & 59.89 & 62.73 & 68.61 \\
\rowcolor{blue!8}
+ FiVeD-unified (Qwen) & 73.36 & 76.04 & 74.67 & 66.49 & 69.28 & 67.82 & 71.61 & 76.93 & 74.17 & 66.76 & 60.78 & 63.60 & 70.07 \\
\quad w/o CSF & 72.63 & 73.22 & 72.89 & 64.61 & 71.59 & 67.88 & 69.22 & 74.98 & 71.98 & 63.05 & 63.33 & 63.15 & 68.98 \\
\quad w/o CSF \& RG & 71.95 & 73.50 & 72.70 & 63.34 & 72.04 & 67.39 & 69.11 & 74.71 & 71.79 & 62.44 & 62.92 & 62.65 & 68.63 \\
\quad w/o CSF \& ETC \& RG & 71.89 & 73.78 & 72.81 & 61.25 & 71.46 & 65.95 & 69.05 & 73.97 & 71.43 & 61.30 & 63.33 & 62.28 & 68.12 \\
\rowcolor{blue!8}
+ FiVeD-unified (DeepSeek-Reasoner) & 74.15 & 74.79 & 74.46 & 66.95 & 69.24 & 68.03 & 71.66 & 77.12 & 74.29 & 66.42 & 62.51 & 64.38 & 70.29 \\
\quad w/o CSF & 72.54 & 73.14 & 72.83 & 63.48 & 71.83 & 67.36 & 69.13 & 75.06 & 71.97 & 63.37 & 63.48 & 63.41 & 68.89 \\
\quad w/o CSF \& RG & 72.44 & 72.15 & 72.27 & 62.09 & 72.21 & 66.76 & 69.15 & 74.01 & 71.48 & 64.07 & 61.52 & 62.74 & 68.31 \\
\quad w/o CSF \& ETC \& RG & 71.93 & 73.32 & 72.60 & 61.66 & 72.04 & 66.45 & 68.55 & 73.77 & 71.06 & 62.72 & 62.59 & 62.65 & 68.19 \\

\midrule
Paraphrase \cite{emnlp/ZhangD0YBL21}  & 71.03 & 66.98 & 70.12 & 65.48 & 57.23 & 61.85 & 68.97 & 68.09 & 68.85 & 64.14 & 56.06 & 60.41 & 65.31 \\
\rowcolor{blue!8}
+ FiVeD-unified (Qwen) & 72.55 & 72.96 & 72.74 & 64.05 & 68.29 & 66.06 & 73.66 & 75.80 & 74.70 & 66.47 & 57.68 & 61.75 & 68.81 \\
\quad w/o CSF & 67.69 & 75.70 & 71.47 & 58.36 & 70.10 & 63.66 & 67.48 & 77.98 & 72.34 & 62.78 & 59.15 & 60.86 & 67.08 \\
\quad w/o CSF \& RG & 67.41 & 75.05 & 71.02 & 56.54 & 71.34 & 63.06 & 67.99 & 76.30 & 71.88 & 59.38 & 60.29 & 59.80 & 66.44 \\
\quad w/o CSF \& ETC \& RG & 66.63 & 75.19 & 70.65 & 55.88 & 70.35 & 62.26 & 66.37 & 76.69 & 71.15 & 58.29 & 60.04 & 59.14 & 65.80 \\
\rowcolor{blue!8}
+ FiVeD-unified (DeepSeek-Reasoner) & 72.83 & 72.82 & 72.82 & 64.37 & 67.46 & 65.82 & 74.17 & 74.79 & 74.47 & 67.33 & 57.53 & 62.04 & 68.79 \\
\quad w/o CSF & 67.05 & 76.19 & 71.32 & 57.27 & 71.42 & 63.53 & 67.40 & 77.86 & 72.25 & 61.83 & 59.74 & 60.75 & 66.96 \\
\quad w/o CSF \& RG & 67.47 & 75.21 & 71.12 & 56.85 & 70.10 & 62.75 & 67.15 & 77.43 & 71.92 & 59.80 & 59.67 & 59.67 & 66.37 \\
\quad w/o CSF \& ETC \& RG & 67.33 & 75.19 & 71.03 & 54.66 & 71.50 & 61.95 & 66.12 & 77.70 & 71.44 & 59.31 & 60.11 & 59.70 & 66.03 \\

\midrule
GAS \cite{acl/Zhang0DBL20} & 65.13 & 72.94 & 68.69 & 57.48 & 62.97 & 60.05 & 65.41 & 68.95 & 67.13 & 59.89 & 56.43 & 58.03 & 63.48 \\
\rowcolor{blue!8}
+ FiVeD-unified (Qwen) & 66.78 & 76.68 & 71.38 & 60.98 & 69.98 & 65.13 & 66.23 & 74.47 & 70.01 & 58.82 & 63.35 & 60.88 & 66.85 \\
\quad w/o CSF & 59.12 & 84.47 & 69.38 & 55.93 & 76.87 & 64.69 & 58.76 & 84.67 & 69.36 & 55.26 & 70.20 & 61.73 & 66.29 \\
\quad w/o CSF \& RG & 57.11 & 84.47 & 68.07 & 51.66 & 78.93 & 62.32 & 59.02 & 80.43 & 67.97 & 50.13 & 73.74 & 59.52 & 64.47 \\
\quad w/o CSF \& ETC \& RG & 57.26 & 83.01 & 67.72 & 50.45 & 77.65 & 61.12 & 58.38 & 81.25 & 67.84 & 49.53 & 73.52 & 59.09 & 63.94 \\
\rowcolor{blue!8}
+ FiVeD-unified (DeepSeek-Reasoner) & 66.41 & 77.23 & 71.40 & 59.74 & 72.37 & 65.39 & 65.17 & 76.26 & 70.25 & 58.33 & 64.16 & 60.99 & 67.01 \\
\quad w/o CSF & 57.88 & 85.62 & 68.93 & 53.85 & 78.76 & 63.89 & 60.08 & 82.72 & 69.57 & 52.44 & 73.85 & 61.18 & 65.89 \\
\quad w/o CSF \& RG & 56.83 & 85.32 & 68.17 & 51.75 & 77.81 & 62.07 & 59.67 & 80.74 & 68.50 & 51.54 & 71.42 & 59.63 & 64.59 \\
\quad w/o CSF \& ETC \& RG & 55.94 & 84.93 & 67.41 & 50.53 & 78.35 & 61.40 & 56.70 & 83.77 & 67.59 & 49.87 & 73.96 & 59.51 & 63.98 \\

\bottomrule
\end{tabular}
\caption{Ablation study of our method on three generative ASTE baselines. P, R, and F1 denote precision, recall, and F1 score, respectively, and Avg. denotes the average F1 across four datasets. CSF, ETC, and RG denote candidate support filtering, error type classification, and rationale generation, respectively.}
\label{tab:ablation}
\end{table*}

\subsection{Ablation Study}
\label{sec:ablation}
Table~\ref{tab:ablation} presents the ablation results of our method on three generative baselines. We evaluate the contribution of three key components: candidate support filtering (CSF), error type classification (ETC), and rationale generation (RG). CSF filters candidates whose support count is lower than \(c_{\min}\). ETC provides fine-grained error-type supervision to help the verifier distinguish different counterfactual errors. RG introduces diagnostic rationale generation as an auxiliary task to encourage more informative verification representations.

Removing CSF consistently decreases the average F1 score across all base generators and diagnostic reasoners. The performance drop is especially clear for Paraphrase and GAS, where the removal of CSF leads to substantially lower precision. This indicates that diverse decoding with an increasing number of candidates can introduce many noisy candidates, and support-based filtering is important for suppressing low-quality triplets before verifier scoring.
Further removing RG causes additional degradation in most settings, showing that rationale generation provides useful auxiliary supervision. By requiring the verifier to generate diagnostic explanations, RG encourages the model to capture why a candidate triplet is correct or incorrect, instead of only learning a binary decision boundary. This improves the robustness of the verification process, particularly for cases involving boundary errors and sentiment polarity errors.
The full removal of CSF, ETC, and RG generally yields the weakest performance, confirming that the proposed components are complementary. ETC contributes fine-grained discriminative signals, RG improves diagnostic reasoning ability, and CSF reduces candidate noise at inference time. Together, these components enable our method to achieve stable improvements over the original generative baselines.

Comparing Qwen2.5-72B-Instruct and DeepSeek-Reasoner, both diagnostic reasoners lead to consistent improvements when used to construct the verification corpus. DeepSeek-Reasoner slightly outperforms Qwen in several average F1 results, while Qwen remains competitive. This suggests that our method is not tied to a specific LLM reasoner and can benefit from different sources of diagnostic supervision.

\begin{figure}
\centering
\includegraphics[width=\linewidth]{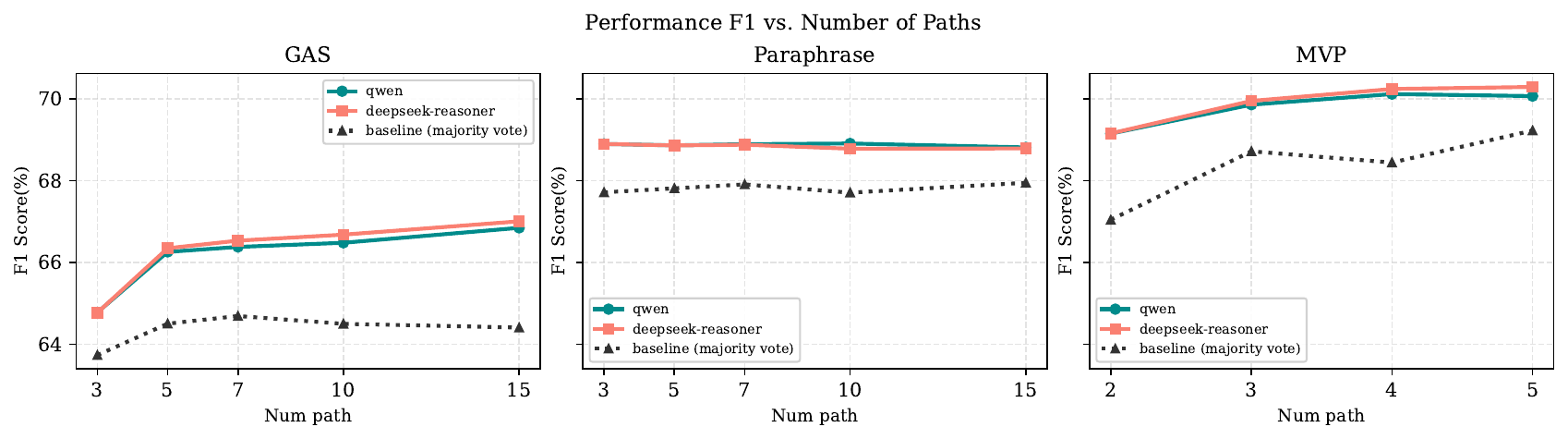}
\caption{F1 performance under different numbers of verification paths across three generative ASTE pipelines. The dotted line denotes the majority-vote baseline, while solid lines denote FiVeD with rationales generated by Qwen and DeepSeek-Reasoner.}
\label{fig:path-trend}
\end{figure}

\subsection{Effect of Verification Paths}
Figure~\ref{fig:path-trend} shows the F1 performance of FiVeD under different numbers of verification paths across three generative ASTE pipelines. 
We compare two rationale generators, Qwen and DeepSeek-Reasoner, and additionally report a majority-vote baseline over the same candidate paths.
Overall, FiVeD consistently outperforms the majority-vote baseline across the three pipelines. 
This confirms that simply aggregating multiple generated candidates by frequency is insufficient for reliable ASTE prediction, because frequent candidates may still contain boundary errors or incorrect sentiment elements. 
In contrast, FiVeD employs the verifier to evaluate the validity of each candidate path and therefore can select candidates based on semantic consistency and structural correctness rather than frequency agreement.
The benefit of increasing the number of paths varies across pipelines. 
For GAS, the F1 score improves noticeably when the number of paths increases from 3 to 5, and then continues to increase more gradually. 
This suggests that additional paths provide useful candidate diversity, especially when the initial candidate set is small. 
For Paraphrase, the performance of FiVeD is relatively stable across different path numbers, indicating that this pipeline already produces comparatively stable candidates and that FiVeD can maintain robust performance without requiring a large sampling budget. 
For MvP, the performance improves as the number of paths increases, showing that verification is particularly useful when multiple candidate orderings or views are available.

We also observe that Qwen and DeepSeek-Reasoner produce similar trends, with DeepSeek-Reasoner slightly outperforming Qwen in several settings. 
This indicates that FiVeD is not tied to a specific rationale generator. 
Instead, it can benefit from different LLM-generated rationales as long as they provide useful semantic evidence for verification.

\begin{figure}
\centering
\includegraphics[width=\linewidth]{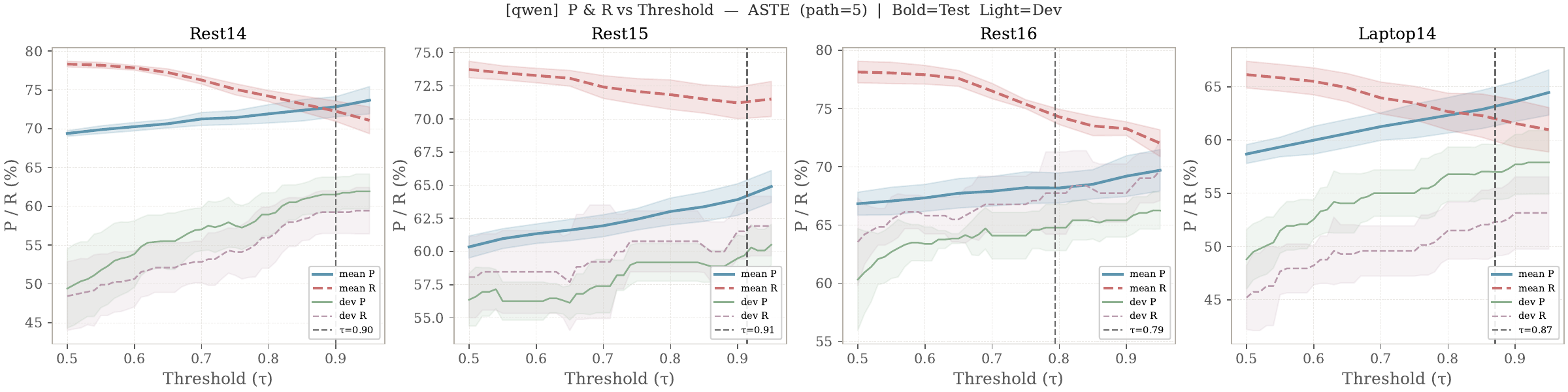}
\caption{Precision-recall curves on the test set for the unified verifier applied to the MVP method with \(5\) candidate paths. The dashed line denotes the quality score threshold \(\tau\) selected on the development set. In this setting, the generative LLM is Qwen.}
\label{fig:threshold-path5-qwen}
\end{figure}

\begin{figure}
\centering
\includegraphics[width=\linewidth]{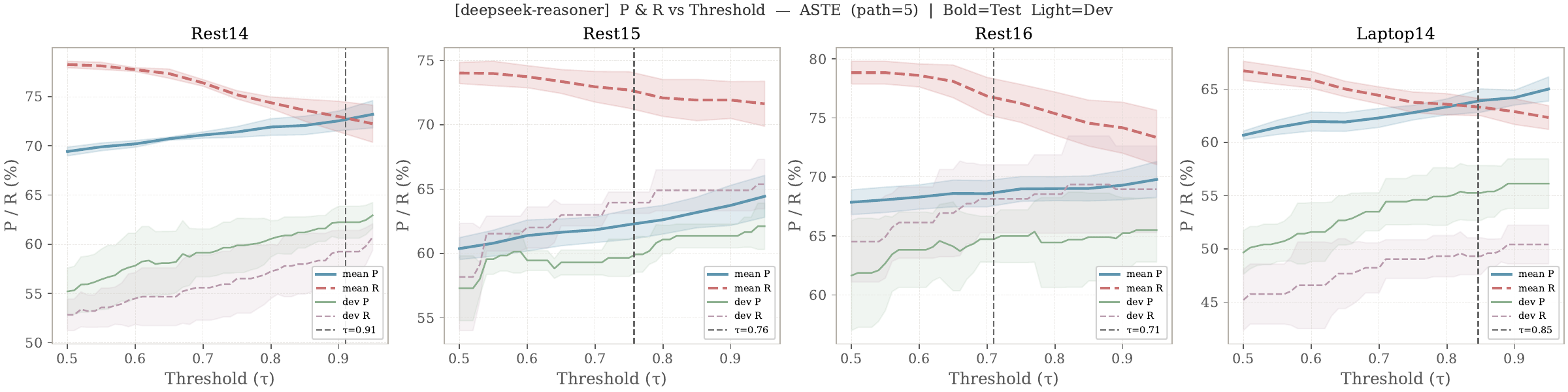}
\caption{Precision-recall curves on the test set for the unified verifier applied to the MVP method with \(5\) candidate paths. The dashed line denotes the quality score threshold \(\tau\) selected on the development set. In this setting, the generative LLM is DeepSeek-Reasoner.}
\label{fig:threshold-path5-deepseek}
\end{figure}

\begin{figure}
\centering
\includegraphics[width=\linewidth]{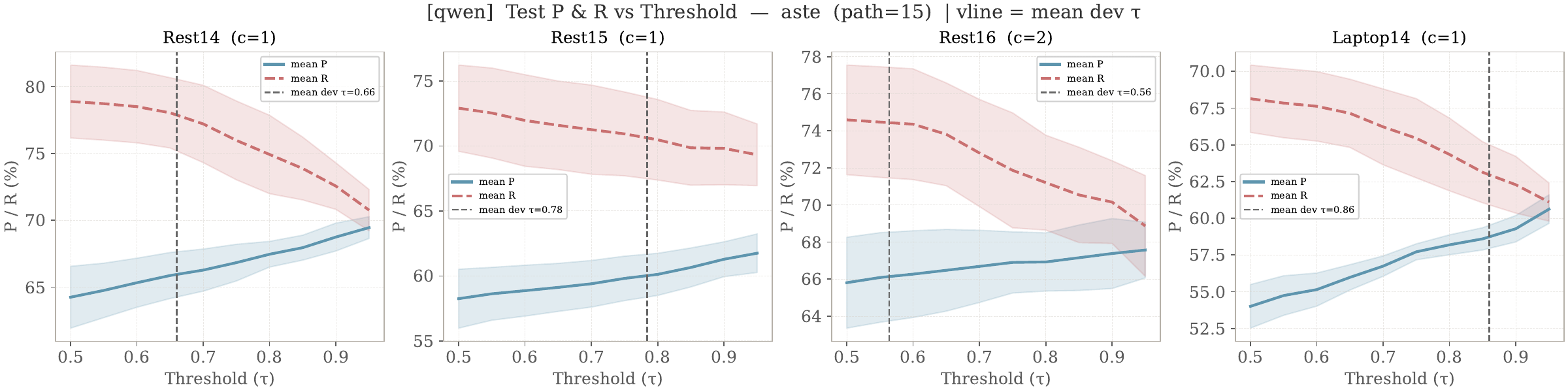}
\caption{Precision-recall curves on the test set for the unified verifier applied to the GAS method with \(15\) candidate paths. The dashed line denotes the quality score threshold \(\tau\), while \(c\) denotes the candidate support filtering threshold, both selected on the development set. In this setting, the generative LLM is Qwen.}
\label{fig:GAS_threshold-path15-qwen}
\end{figure}

\begin{figure}
\centering
\includegraphics[width=\linewidth]{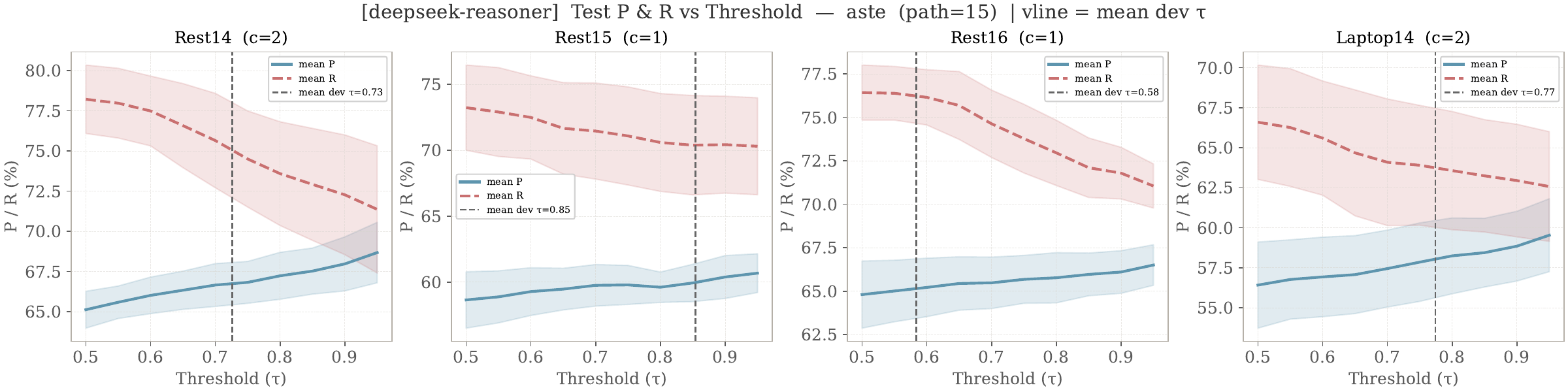}
\caption{Precision-recall curves on the test set for the unified verifier applied to the GAS method with \(15\) candidate paths. The dashed line denotes the quality score threshold \(\tau\), while \(c\) denotes the candidate support filtering threshold, both selected on the development set. In this setting, the generative LLM is DeepSeek-Reasoner.}

\label{fig:GAS_threshold-path15-deepseek}
\end{figure}

\begin{figure}
\centering
\includegraphics[width=\linewidth]{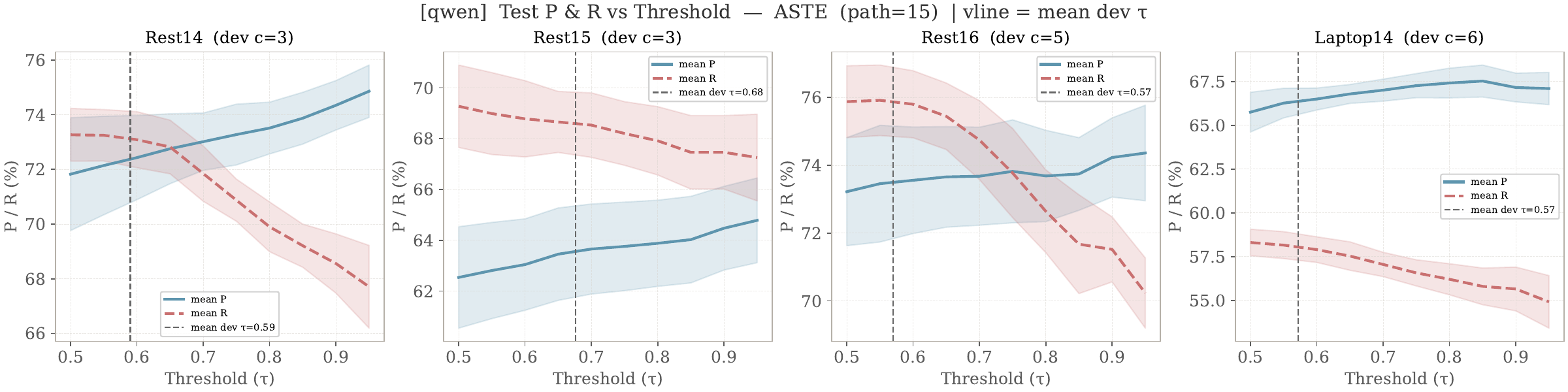}
\caption{Precision-recall curves on the dev and test set for the unified verifier applied to the Paraphrase method with \(15\) candidate paths. The dashed line denotes the quality score threshold \(\tau\), while \(c\) denotes the candidate support filtering threshold, both selected on the development set. In this setting, the generative LLM is Qwen.}
\label{fig:PR_threshold-path15-qwen}
\end{figure}

\begin{figure}
\centering
\includegraphics[width=\linewidth]{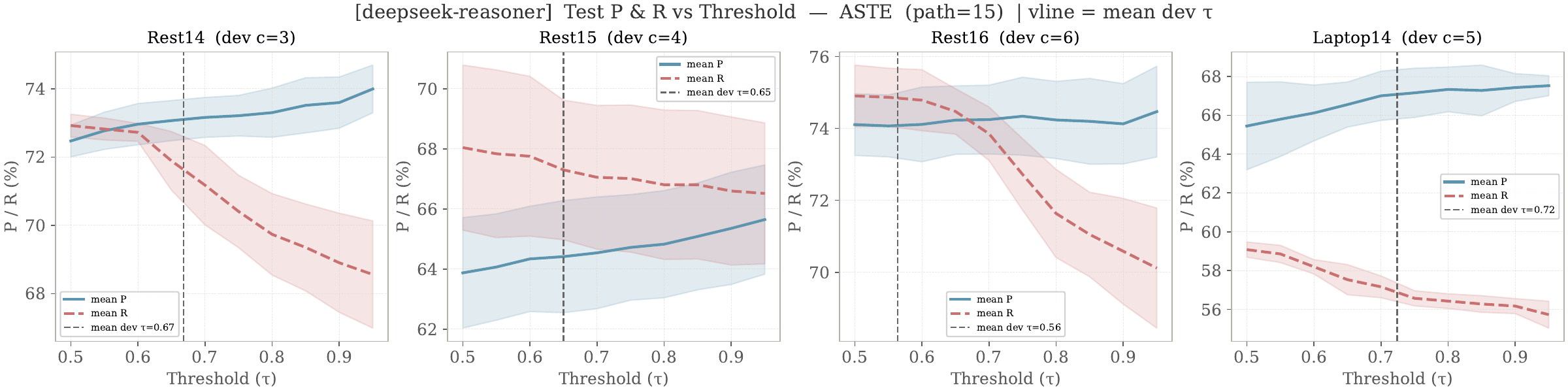}
\caption{Precision-recall curves on the dev and test set for the unified verifier applied to the Paraphrase method with \(15\) candidate paths. The dashed line denotes the quality score threshold \(\tau\), while \(c\) denotes the candidate support filtering threshold, both selected on the development set. In this setting, the generative LLM is DeepSeek-Reasoner.}
\label{fig:PR_threshold-path15-deepseek}
\end{figure}

\subsection{Verification Threshold Analysis}
\label{sec:verifier-behavior}

\label{sec:threshold-analysis}

FiVeD uses a verification threshold \(\tau\) to decide whether a candidate should be accepted. 
The choice of \(\tau\) controls the trade-off between precision and recall: a larger threshold makes the verifier more conservative, which usually improves precision but reduces recall; a smaller threshold accepts more candidates, which can improve recall but may introduce more false positives.

Figure~\ref{fig:threshold-path5-qwen} and Figure~\ref{fig:threshold-path5-deepseek} present the precision-recall curves of MvP when using two to five verification paths. 
The maximum number of views is limited to five by the element-order design adopted during MVP training for ASTE.
The light curves correspond to validation set precision and recall, while the bold curves correspond to test set precision and recall. 
The vertical dashed line denotes the threshold selected on the validation set. 
Across datasets, precision generally increases as \(\tau\) becomes larger, while recall decreases. 
This expected trade-off indicates that the verifier assigns meaningful quality scores, where candidates with higher scores are more likely to be valid and can therefore be retained under stricter thresholds. 
Moreover, the threshold selected on the development set transfers reasonably well to the test set, suggesting that the learned scores are not merely overfitted to a particular split but provide a stable basis for candidate selection. 
These results show that, by incorporating quality score supervision, FiVeD goes beyond binary validity checking and offers an adjustable mechanism for balancing precision and recall.

We further examine whether this behavior holds under a larger candidate set and across different generative ASTE pipelines. 
Figure~\ref{fig:GAS_threshold-path15-qwen} and Figure~\ref{fig:GAS_threshold-path15-deepseek} report the results for GAS with up to \(15\) verification paths, while Figure~\ref{fig:PR_threshold-path15-qwen} and Figure~\ref{fig:PR_threshold-path15-deepseek} report the results for Paraphrase. 
The observed trends are consistent with those of MvP that increasing \(\tau\) leads to more conservative candidate selection, and the precision-recall trade-off remains stable. 
This suggests that quality scores predicted by FiVeD remain informative even when more candidate paths are introduced and when FiVeD is applied to different generative baselines. 
In practice, a higher threshold can be used in scenarios that prioritize correctness, whereas a lower threshold can be selected when broader extraction coverage is desired.
\section{Discussion}
\label{sec:discussion}

The preceding experiments demonstrate the overall effectiveness of FiVeD in ASTE verification.
In this section, we move beyond aggregate metrics and analyze the supervision signals that guide the verifier. Specifically, we investigate how different task weights influence the learning of fine-grained verification objectives, and examine whether the quality scores used as supervision reflect the severity of different error types. These analyses help explain how FiVeD leverages diagnostic LLM supervision to learn fine-grained and reliable verification decisions. We conclude the section by discussing promising directions for future research.

\begin{figure}[h]
\centering
\includegraphics[width=0.95\linewidth]{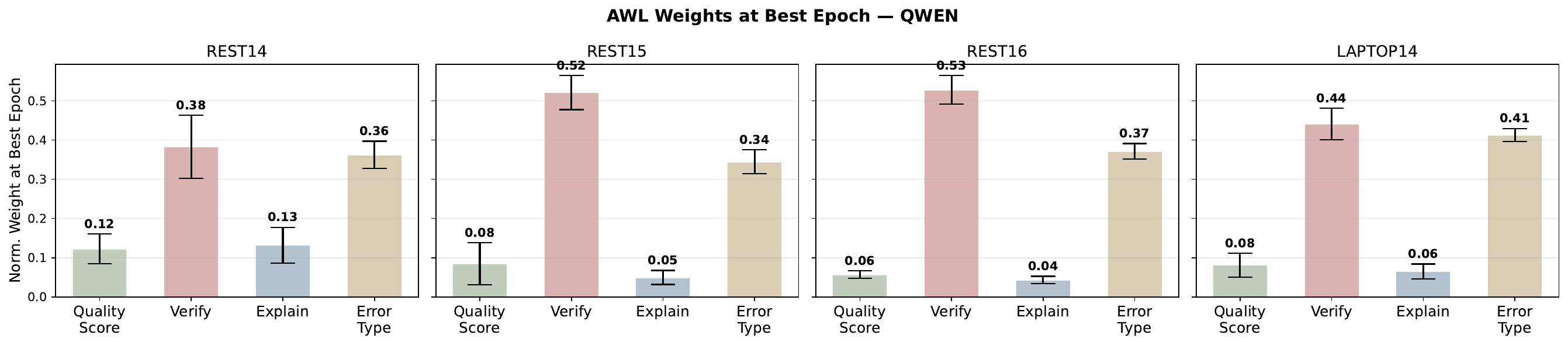}
\caption{The automatically learned loss weight assigned to each sub-task with rationale generation from Qwen.}
\label{fig:awl-qwen}
\end{figure}

\begin{figure}[h]
\centering
\includegraphics[width=0.95\linewidth]{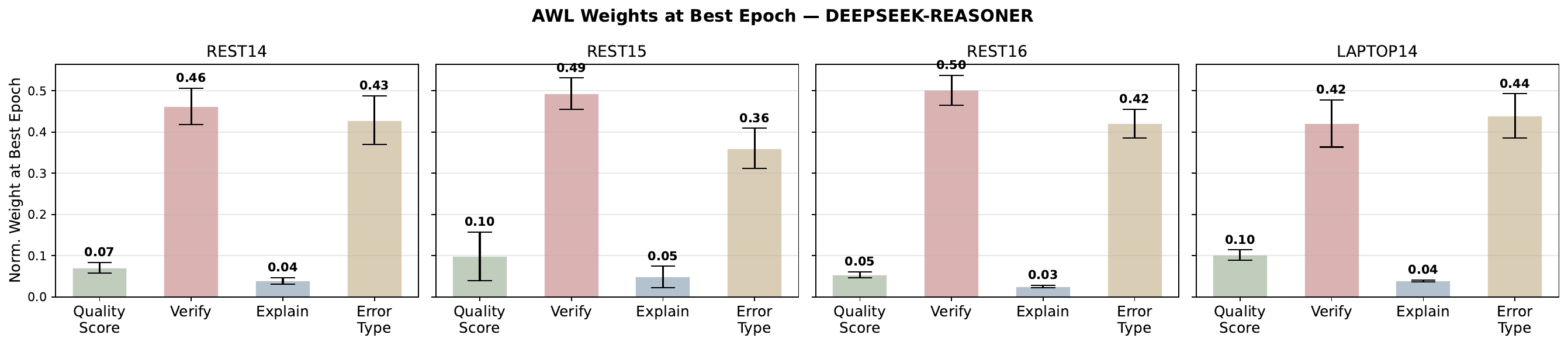}
\caption{The automatically learned loss weight assigned to each sub-task with rationale generation from DeepSeek-Reasoner.}
\label{fig:awl-deepseek}
\end{figure}

\subsection{Analysis of Task Weights}
\label{sec:awl-analysis}

FiVeD is trained with multiple objectives, including quality score prediction, validity verification, rationale generation, and error type classification. 
To balance these objectives, we adopt automatically learned loss weighting. 
Figure~\ref{fig:awl-qwen} and Figure~\ref{fig:awl-deepseek} report the normalized task weights at the best validation epoch when LLMs use Qwen and DeepSeek-Reasoner.
Across datasets and reasoning generators, validity verification and error type classification receive relatively larger weights. 
This pattern suggests that the automatically learned weighting scheme places greater emphasis on objectives that provide direct discriminative signals for candidate selection. 
At the same time, the smaller weights assigned to quality score prediction and explanation generation indicate that these objectives play more complementary roles during optimization.
Although its learned weight is smaller, the quality score objective provides a continuous ranking signal that complements the discrete validity label. 
Together, quality score prediction and validity verification allow FiVeD to distinguish not only correct and incorrect candidates but also different levels of candidate quality. 
Similarly, the explanation objective receives the smallest weight, but it still contributes by encouraging the verifier to align its decisions with diagnostic reasoning. 
The consistently low explanation weight also helps prevent noisy or overly verbose rationales from overwhelming the core verification objective.

These learned weights are highly consistent between Qwen and DeepSeek-Reasoner, indicating stable training dynamics across different reasoning generators. 
Overall, the weight patterns reveal a division of labor among the objectives, where verification and error type classification provide strong discriminative signals, quality score prediction provides a continuous ranking signal, and explanation generation provides diagnostic reasoning guidance. 
Consistent with this interpretation, the ablation results in Section \ref{sec:ablation} show that all objectives contribute to the final performance and provide complementary supervision during training.

\begin{figure}[h]
\centering
\includegraphics[width=0.95\linewidth]{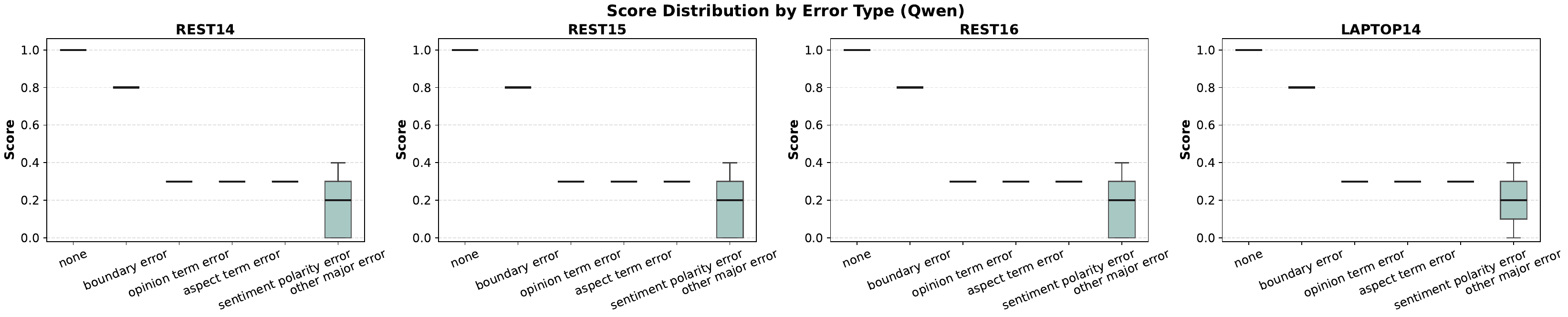}
\caption{Distribution of constructed quality scores by error type when using Qwen to generate diagnostic reasoning supervision.}
\label{fig:error-type-qwen}
\end{figure}

\begin{figure}[h]
\centering
\includegraphics[width=0.95\linewidth]{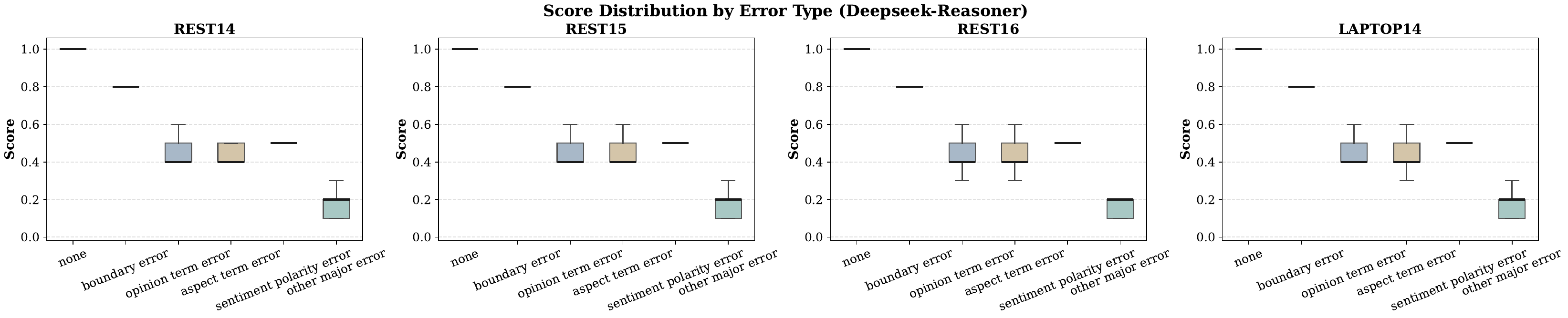}
\caption{Distribution of constructed quality scores by error type when using DeepSeek-Reasoner to generate diagnostic reasoning supervision.}
\label{fig:error-type-deepseek}
\end{figure}

\subsection{Quality Scores Distribution by Error Type}
\label{sec:error-type-analysis}

To examine whether the constructed supervision provides informative training signals, we analyze the distribution of quality scores across different error types. 
Figure~\ref{fig:error-type-qwen} and Figure~\ref{fig:error-type-deepseek} show the distributions of constructed quality scores when diagnostic reasoning supervision is assisted by Qwen and DeepSeek-Reasoner, respectively. 
In both settings, correct predictions, denoted as \textit{none}, are assigned the highest scores and are concentrated near the top of the score range. 
This indicates that the constructed supervision clearly separates valid candidates from erroneous ones. 
In contrast, candidates with major errors receive substantially lower scores. 
For example, \textit{other major error} consistently receives the lowest scores across datasets, indicating that the supervision signal penalizes candidates that deviate substantially from the gold triplet structure.

The score distributions further reflect different degrees of error severity. 
Boundary errors receive relatively high scores compared with other error types, because a boundary mismatch may still preserve the core aspect-opinion-sentiment relation. 
Aspect term and opinion term errors receive lower scores, since replacing either term changes the semantic target or the supporting opinion expression. 
Sentiment polarity errors are also strongly penalized, as they directly change the sentiment understanding of the extracted triplet. 
These patterns suggest that the constructed quality scores encode a hierarchy of error severity rather than providing only binary correctness labels.

Comparing the two generators, DeepSeek-Reasoner tends to produce more gradual score variations across error types, whereas Qwen exhibits a more discretized scoring pattern with scores concentrated around a few levels.
Despite these differences, both LLMs yield the same ordering by severity, where correct predictions receive the highest scores, minor errors receive intermediate scores, and major errors receive the lowest scores. 
This consistency suggests that the constructed diagnostic supervision encodes structured severity information, providing a basis for FiVeD to learn a more nuanced verification function that captures different degrees of error severity rather than only binary validity discrimination.

\subsection{Future Directions}
Future work can extend our framework in several directions. One direction is to develop more adaptive error taxonomies that account for domain or dataset-specific error patterns while retaining the compact structure used in this work. Another direction is to incorporate human feedback into the supervision construction, enabling interactive refinement of LLM-generated quality scores and rationales. Besides, it would also be valuable to apply the verifier to broader ASTE settings, such as multilingual ASTE, where the same verification paradigm may help analyze diverse candidate error patterns. Finally, jointly optimizing candidate generation and verification may further enhance extractor performance.
\section{Related Work}
\label{sec:relatedwork}
\subsection{Aspect Sentiment Triplet Extraction}

Aspect Sentiment Triplet Extraction (ASTE) is a fine-grained sentiment analysis task 
that extracts aspect terms, opinion terms, and the sentiment polarities expressed toward 
the aspects as structured triplets \cite{tkde/SchoutenF16, zhang-etal-2024-sentiment}. 
Compared with subtasks such as aspect term extraction, opinion term extraction, aspect-level sentiment classification, and aspect-opinion pair extraction, ASTE requires models to jointly identify all sentiment elements and determine 
their semantic associations. A predicted triplet is therefore valid only when 
its three components are correctly identified and mutually consistent in context 
\cite{lai2025E4L, bodke-etal-2025-pastel}.

Existing studies have primarily focused on improving end-to-end extraction models. 
Early work formulates ASTE as a sequence labeling problem with specially designed tagging schemes \cite{Peng_Xu_Bing_Huang_Lu_Si_2020, yan-etal-2021-unified, xianlong-etal-2023-tagging}. To better capture aspect-opinion interactions, subsequent studies adopt formulations that explicitly model their structural dependencies, such as span-based extraction \cite{xu-etal-2021-learning, zhao-etal-2024-dual, EPMEI}, table filling \cite{SIMSTAR, IJCAI-TT-TABLE}, grid tagging \cite{sun-etal-2024-minicongts}, and graph-based modeling \cite{chen-etal-2022-enhanced}. Another line of work formulates ASTE as a machine reading comprehension (MRC) task \cite{Dual-MRC21, zhai-etal-2022-com, Zou2024AMS}. Instead of extracting a triplet in a single pass, these methods decompose ASTE into several query-conditioned prediction steps and assemble the final triplets during inference. More recently, generative methods have cast ASTE as a text-to-text problem, where pretrained sequence-to-sequence models generate triplets in a structured format \cite{emnlp/ZhangD0YBL21, acl/Zhang0DBL20}. These approaches benefit from the strong representation and generation abilities of pretrained language models and can flexibly produce variable-length structured outputs. Prompt-based and instruction-tuned methods further exploit the knowledge and generalization capabilities of large pretrained models for structured sentiment extraction \cite{acl/MvP}.

Despite this progress, most methods optimize the extractor itself, and decoded triplets are typically treated as final outputs. In practice, ASTE predictions may still contain subtle errors, such as incorrect boundaries, mismatched element terms, or partially correct triplets, which are difficult to eliminate solely through extractor design. 
A recent attempt, PASTEL \cite{bodke-etal-2025-pastel}, decomposes ASTE into structured subtasks and applies an LLM-as-judge strategy to evaluate candidate triplets composed from the subtask outputs. While this introduces an additional verification signal, directly using off-the-shelf LLMs as judges remains unstable for element-level ASTE checking, since LLMs themselves can struggle with structured sentiment extraction and produce hallucinated or inconsistent 
judgments \cite{lai2025llmsteamup, zhang-etal-2024-sentiment}.

\subsection{Verification for Structured Prediction}

In the era of large language models, detecting and mitigating hallucinations has become increasingly important \cite{lightman2024lets, lai2024mtisa}. Verification improves prediction trustworthiness by assessing whether model outputs are faithful to the input, internally consistent, or supported by external evidence \cite{Li2024ASO, rvisa, zhu-etal-2023-solving}. Beyond post-hoc checking, it has been instantiated in many forms, including confidence estimation and calibration, candidate filtering, reranking, verifier-guided decoding, and reward modeling for reinforcement learning with verifiable rewards \cite{Zhang2025100DA, xu-etal-2024-sayself}.

Verification is especially valuable for structured prediction, where outputs may be well-formed yet semantically incorrect or only partially supported by the input. Calibration methods seek to align prediction scores with actual correctness, while reranking and verifier-based approaches evaluate candidate structures and select more stable outputs. Related mechanisms have also been 
studied in reasoning and puzzle-solving, where verifiers check format validity, 
constraint preservation, and final-answer correctness \cite{xie2025logicrlunleashingllmreasoning, lin2025zebralogic}, and in application-oriented tasks such as software repair, retrieval-augmented question answering, and chemistry-oriented generation, where verification signals assess task-specific correctness, evidence support, 
or syntactic validity \cite{wei2026swerl, Huang_2026, Jha_2025}. 
A common observation across these studies is that output correctness often decomposes into multiple dimensions rather than a single binary judgment.

This perspective applies naturally to ASTE, where errors may arise from 
individual sentiment elements, their boundaries, sentiment labels, or 
aspect-opinion associations. However, existing ASTE systems typically lack 
a dedicated verifier, and as discussed above, direct LLM-as-judge validation 
can be unstable and hard to control. Motivated by this gap, we propose 
\textbf{FiVeD}, a fine-grained verification framework for ASTE that uses diagnostic reasoning from LLMs as auxiliary supervision rather than as 
the final judgment. By converting LLM reasoning into structured training 
signals for a dedicated verifier, FiVeD provides diagnostic assessment while reducing dependence on direct LLM decisions at inference time.
\section{Conclusion}
\label{sec:conclusion}
In this paper, we propose FiVeD, a fine-grained verification framework with diagnostic reasoning for aspect sentiment triplet extraction. FiVeD addresses the challenging problem of triplet verification by assessing the validity and quality of candidate triplets after extraction. Through multi-objective supervision, including validity classification, quality score estimation, error type classification, and rationale generation, FiVeD enables more fine-grained and interpretable verification by encouraging the verifier to jointly assess triplet validity and diagnose potential error sources.
Experiments across multiple ASTE baselines show that FiVeD consistently improves extraction performance for base extractors, enables controllable precision-recall trade-offs, and achieves superior results. These findings demonstrate the effectiveness of FiVeD as a flexible plug-and-play verification module. More broadly, they highlight the value of fine-grained verification for enhancing the reliability of ASTE systems and suggest its potential applicability to broader sentiment analysis and structured information extraction tasks.


\begin{acks}
This work was supported by the Research Impact Fund by the Research Grants Council of Hong Kong (Project No. 130272) and a grant from the Research Grants Council of the Hong Kong Special Administrative Region, China (R1015-23); the Faculty Research Grants (SDS24A8, SDS25A15 and SDS24A19), the Interdisciplinary \& Strategic Research Grant (ISRG252606), and the Direct Grants (DR25E8 and DR26F2) of Lingnan University, Hong Kong.
\end{acks}

\bibliographystyle{ACM-Reference-Format}
\bibliography{main}

\appendix

\end{document}